\documentclass[10pt,twocolumn,letterpaper]{article}

\usepackage{cvpr}              
\definecolor{cvprblue}{rgb}{0.21,0.49,0.74}
\usepackage[pagebackref,breaklinks,colorlinks,allcolors=cvprblue]{hyperref}

\usepackage{multirow}
\usepackage{amsmath}
\usepackage{amssymb}
\usepackage{algorithm}
\usepackage{algorithmicx}
\usepackage{algpseudocode}
\usepackage{tabu}
\usepackage{siunitx}
\usepackage{adjustbox}
\usepackage{siunitx}
\usepackage{makecell}
\usepackage{xcolor}
\usepackage{colortbl}
\usepackage{color}
\usepackage{bm}
\usepackage{multirow}
\usepackage{blindtext}
\usepackage{comment}
\usepackage{bbm}
\usepackage{booktabs}
\usepackage{wrapfig}
\usepackage{amsfonts}
\usepackage{graphicx}
\usepackage[accsupp]{axessibility}  
\def\paperID{} 
\def\confName{CVPR}
\def\confYear{2026}

\title{Token Reduction via Local and Global Contexts Optimization for Efficient Video Large Language Models}

\author{Jinlong Li$^{1,\dagger}$ ~~~~~~~ {Liyuan Jiang}$^{2}$ ~~~~~~~ {Haonan Zhang}$^{3}$  ~~~~~~~  {Nicu Sebe$^{1}$} \\
$^1$ University of Trento \\ $^2$ Tsinghua University \\ $^3$ University of Electronic Science and Technology of China
}

\begin{document}
\maketitle

\newcommand\blfootnote[1]{%
\begingroup 
\renewcommand\thefootnote{}\footnote{#1}%
\addtocounter{footnote}{-1}%
\endgroup 
}
    {
        \blfootnote{
          $^\dagger$Corresponding author: tyronejinlongli@gmail.com.
    
    }
}

\begin{abstract}

Video Large Language Models (VLLMs) demonstrate strong video understanding but suffer from inefficiency due to redundant visual tokens. Existing pruning primary targets intra-frame spatial redundancy or prunes inside the LLM with shallow-layer overhead, yielding suboptimal spatiotemporal reduction and underutilizing long-context compressibility. All of them often discard subtle yet informative context from merged or pruned tokens. In this paper, we propose a new perspective that elaborates token \textit{\textbf{A}nchors} within intra-frame and inter-frame to comprehensively aggregate the informative contexts via local-global \textit{\textbf{O}ptimal} \textit{\textbf{T}ransport} (\textbf{AOT}). Specifically, we first establish local- and global-aware token anchors within each frame under the attention guidance, which then optimal transport aggregates the informative contexts from pruned tokens, constructing intra-frame token anchors. Then, building on the temporal frame clips, the first frame within each clip will be considered as the keyframe anchors to ensemble similar information from consecutive frames through optimal transport, while keeping distinct tokens to represent temporal dynamics, leading to efficient token reduction in a training-free manner. Extensive evaluations show that our proposed AOT obtains competitive performances across various short- and long-video benchmarks on leading video LLMs, obtaining substantial computational efficiency while preserving temporal and visual fidelity. Project webpage: \href{https://tyroneli.github.io/AOT}{AOT}.

\end{abstract}    
\section{Introduction}
\label{sec:introdction}

Video Large Language Models (VLLMs)~\cite{bai2025qwen2,zhu2025internvl3,zhang2023video,lin2023video,cheng2024videollama,li2024mvbench,weng2024longvlm,wu2023visual,wang2023chatvideo,li2023videochat,chen2024sharegpt4video} nowadays have showcased remarkable prominence for complex video understanding and comprehension. The visual encoder~\cite{zhai2023sigmoid,radford2021learning} converts sampled frames into video token sequences, which LLM then processes alongside text sequences to generate the responses. Increasingly demandings further require VLLMs to process longer and more complex video scenarios~\cite{li2024llama,chai2024auroracap,team2024gemini,weng2024longvlm}.

Despite their effectiveness, the high computational cost and memory consumption of inference pose significant challenges, particularly making  inference computationally expensive when processing videos with numerous frames, which takes tens of thousands of input token count. Though some approaches~\cite{jin2024chat,maaz2023video,shen2024longvu,xu2024pllava,zohar2025apollo,song2024moviechat} perform trainable compression module to alleviate this issue, they still demand extensive training or fine-tuning, leading to high training costs. While prior works~\cite{yang2025visionzip,xing2024pyramiddrop,tao2025dycoke,bolya2022token,shang2025llava,zhang2024sparsevlm,shao2025holitom} have explored model compression and token pruning to mitigate the efficiency problem, achieving a desirable balance between efficiency and performance, but fail to exploit temporal dependencies across sampled frames.

\begin{figure}[t]
    \centering
    \includegraphics[width=\linewidth]{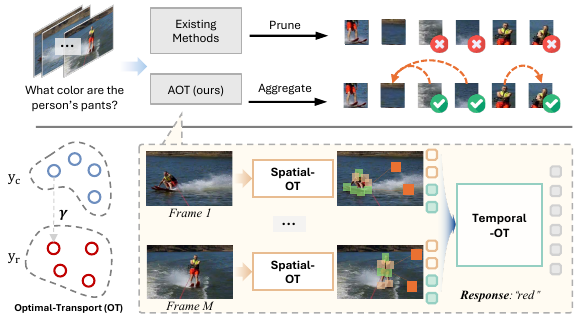}
    \caption{The top is the essential differences compared with common token reduction methods, instead of simply removing unimportant or merging very similar tokens, ours utilizes a global optimization strategy to further exploit and aggregate necessary semantic and context from these onto the remaining tokens. Bottom is our proposed pipeline to adopt Optimal Transport to aggregate information within intra- and inter-frame levels for video tokens.}
    \label{fig:motivation_fig}
    \vspace{-0.2cm}
\end{figure}

Thus, developing effective methods to reduce video token redundancy while preserving critical semantical and contextual information is crucial for the widespread utility of video LLMs. Recently, video compression methods like DyCoke~\cite{tao2025dycoke} and PruneVid~\cite{huang2024prunevid} mainly prune at the LLM prefilling and decoding stages, processing video tokens analogous to textual tokens and ignoring specific characteristics of video. Moreover, some token pruning approaches~\cite{fu2024framefusion,bolya2022token,yang2025visionzip,shang2025llava} take the most-discriminative tokens while simply removing the low-discriminative or fusing many similar ones, vulnerable to token selection and neglecting informative regions or involving noisy backgrounds. A more video-specific pruning approach is necessary to fully exploit spatiotemporal redundancy while preserving appropriate visual contexts.

To alleviate this limitation, in this paper, we propose a new perspective that elaborates token \textit{\textbf{A}nchors} within intra-frame and inter-frame to comprehensively aggregate the semantic and context information via local-global \textit{\textbf{O}ptimal} \textit{\textbf{T}ransport} (\textbf{AOT}). Inspired by the AnyRes technique from MLLMs~\cite{li2024llava,chen2024far,zhang2024llavanextvideo}, we first perform a grid-wise local token selection to maintain the local prior while a global-level selection exclusive will be applied to select the global tokens with attention guidance, together serving as token \textit{anchors} for each frame. This strategy retains semantically important and spatially diverse token candidates.

Building on this insight, how to appropriately measure the relationship between the selected and unselected tokens remains a critical exploration, which aims to abstract necessary information. To address this, we introduce Optimal Transport (OT) to measure the distances between the selected token anchors and unselected tokens through a global optimization strategy. For intra-frame token pruning, we formulate token anchors and unselected tokens as the samplings of two discrete distributions and use OT to encourage fine-grained measurement matching (\textit{few-to-many}), while inverse cosine similarity among token sets serves as the cost matrix to be optimized. 
The distance calculation between token sets will be modeled as a discrete probability distribution where each token has an equal probability value based on optimal transport theory~\cite{villani2008optimal}.
Each token in the unselected set is a supplier who supplies a certain level of necessary contexts, and each token anchor is a demander who needs one unit of context. 
In this sense, each token anchor comprehensively consolidates the aggregation from unselected tokens globally under the optimized transport plan.

For inter-frame token pruning, we further utilize Optimal Transport to tackle temporal redundancy by establishing the first frame within each frame clip as the anchors, and then ensembling similar tokens across the consecutive frames and gradually updating the token anchors, while keeping dissimilar tokens to represent key temporal dynamics. This can be employed with uniform sampling or adaptive clustering frame clip.
After that, AOT leads to necessary local-global semantic and context aggregation across spatiotemporal dimensions by adopting Optimal Transport to assemble informative cues from numerous tokens under merging or removing. Differing from existing methods~\cite{tao2025dycoke,huang2024prunevid,liu2025hybrid,shen2024longvu}, our approach considers detailed intrinsic contributions of these tokens to compact token anchors shown in Fig.~\ref{fig:motivation_fig}, significantly accelerating Video LLM inference while preserving both temporal and visual integrity. After formulation, finding the best assignment solution is converted to solve an optimal transport plan, which can be quickly and efficiently solved by the off-the-shelf Sinkhorn-Knopp Iteration~\cite{cuturi2013sinkhorn} using GPU.

To evaluate the effectiveness of our method, we perform extensive experiments on  LLaVA-OneVision 7B and LLaVA-Video 7B models across MVBench~\cite{li2024mvbench}, LongVideoBench~\cite{wu2024longvideobench}, EgoSchema~\cite{mangalam2023egoschema}, and VideoMME~\cite{fu2025video}. Our approach reduces computational costs to just 8.3\% of the original FLOPs and prunes 90\% of video tokens while remarkably preserving 97.6\% of the original model’s performance across all benchmarks. These results clearly demonstrate the substantial practical advantages of our token reduction framework for efficient video LLM inference.
Our main contributions are:

\begin{itemize}
    \item To the best of our knowledge, we are the first to investigate how to aggregate subtle yet informative semantics and contexts from merging or removing tokens into remaining tokens, instead of simply merging or removing.
    
    \item We study how to first facilitate token anchors that consider both local and global prior, leading to semantically important and spatially diverse candidates.
    
    \item We explore Optimal Transport to aggregate spatiotemporal context from the transport plan within intra- and inter-frame to preserve temporal and visual fidelity with a training-free pipeline.

    \item We evaluate our method in a wide spectrum of video benchmarks and present competitive performances under constrained token budgets.
\end{itemize}

\section{Related Works}
\label{sec:relatedworks}

\subsection{Video Large Language Models}

With the rapid progress of Large Language Models (LMMs)~\cite{achiam2023gpt,chiang2023vicuna,taori2023stanford} and Multimodal Large Language Models (MLLMs)~\cite{alayrac2022flamingo,li2023blip,liu2024improved,liu2024llavanext,team2023gemini,bai2025qwen2,zhu2025internvl3,zhang2025omnicharacter,zhang2025text}, nowadays there has been growing interest in developing Video LLMs (VLLMs)~\cite{bai2025qwen2,zhu2025internvl3,zhang2023video,lin2023video,cheng2024videollama,li2024mvbench,weng2024longvlm,wu2023visual,wang2023chatvideo,li2023videochat,zeng2022video,chen2024sharegpt4video}, enabling video understanding and question answer tasks. They can be categorized into general Video LLMs and Video LLMs with visual token compression during training time. General Video LLMs~\cite{lin2023video,li2024llava,an2025llava,zhang2024video,cheng2024videollama,bai2025qwen2,zhu2025internvl3} extract directly the raw video frames tokens and then simply apply pooling before feeding to the LLM. Moreover, to handle spatiotemporal information for videos, some methods~\cite{bai2025qwen2,zhang2024video} extend the 2D image position encoding into video by inducing a temporal dimension. Video LLMs with visual token compression during the training-time~\cite{xu2024pllava,maaz2023video,jin2024chat,shen2024longvu,zohar2025apollo,li2024llama} propose to reduce video tokens significantly to enable long-context video processing.

However, due to the high spatiotemporal demands of complex video understanding tasks, the visual tokens dominate the following LLM computation overhead since the tokens length can reach up to one million for hours-long videos~\cite{team2024gemini}, substantially increasing inference time and memory consumption. Though some methods~\cite{lin2024vila,liu2025nvila} aim to optimize token utility, they still require model fine-tuning and demand considerable hardware resources. This underscores a critical need for developing more efficient, training-free token compression methods specifically for video LLMs, bypassing the need for costly model adaptations and significant hardware consumption.

\subsection{Image Visual Token Compression}

Token compression has emerged as an effective approach to reduce the computational overhead and complexity in transformer visual encoders and large language models, such as ViT~\cite{dosovitskiy2020image}, CLIP~\cite{radford2021learning} and SigLip~\cite{zhai2023sigmoid}. Pioneering works such as ToMe~\cite{bolya2022token} and FastV~\cite{chen2024image} have explored methods for visual token merging spatially and text-guided pruning to improve the efficiency of LVLMs~\cite{kong2025token}. Hence, this line of approaches can be broadly divided into two main categories: (1) Text-agnostic token compression approaches~\cite{yang2025topv,arif2025hired,shang2025llava,wen2025stop,yang2025visionzip,zhang2024cls,zhang2025vscan}, which discover and merge or remove redundant or uninformative visual tokens during visual encoding. VisionZip~\cite{yang2025visionzip} picks out the dominant visual tokens based on \texttt{[CLS]} attention scores, while LLaVA-PruMerge~\cite{shang2025llava} identifies significant spatial redundancy among visual tokens and applies an adaptive visual token reduction strategy to reduce the number of visual tokens. (2) Text-guided pruning approaches~\cite{tan2025tokencarve,zhang2024sparsevlm,xing2025conical,xing2024pyramiddrop,liu2024multi,ye2025atp,khaki2025sparsevila}, which target at removing visual tokens that are irrelevant to the text query during the LLM decoding phase. SparseVLM~\cite{zhang2024sparsevlm} utilizes an iterative sparsification strategy to exploit visual-relevant text tokens to rank the significance of vision tokens. PyramidDrop~\cite{xing2024pyramiddrop} applies progressive pruning of tokens at different stages within the LLM.

\subsection{Video Visual Token Compression}

Nevertheless, when it comes to temporal dependencies between video frames, specialized compression designs need to be tailored. TempMe~\cite{shen2024tempme} extends progressive spatial tokens reduction by merging neighboring clips to minimize temporal redundancy.
DyCoke~\cite{tao2025dycoke} merges tokens across frames and applies dynamic KV cache reduction. However, its pruning during the prefilling stage struggles to achieve substantial token reduction while maintaining accuracy. PruneVID~\cite{huang2024prunevid} applies both merging and pruning across successive shallow LLM layers, but repeated pruning operations adversely affect overall efficiency. FastVID~\cite{shen2025fastvid} enhances compression by combining temporal segmentation with spatio-temporal token merging. FrameFusion~\cite{fu2024framefusion} applies both merging and pruning across successive shallow LLM layers, but repeated pruning operations adversely affect overall efficiency. Differently, our method proposes to construct compact token anchors across spatiotemporal to aggregate both local- and global-level semantic and contextual information through an optimization strategy.

\section{Methodology}
\label{sec:methodology}

\begin{figure*}[t]
    \centering
    \includegraphics[width=\linewidth]{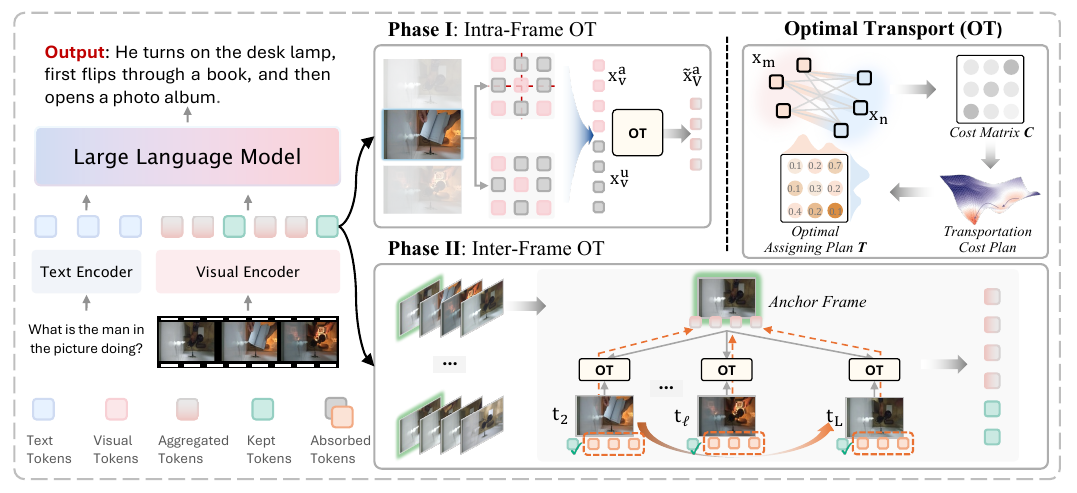}
    \vspace{-0.6cm}
    \caption{Overall pipeline of our \textbf{AOT}. Our method compresses tokens of video LLMs across spatiotemporal through optimal transport, first establishing token anchors within each frame to cover semantically important and spatially diverse token candidates, then utilizing optimal transport to aggregate the necessary informative cues within Intra-Frame at phase I, and finally shifting the optimization strategy into temporal within Inter-Frame at phase II. The proposed AOT preserves both temporal and visual integrity by utilizing efficient Sinkhorn-Knopp Iteration to solve the optimal transport plan assignment.}
\label{fig:method}
\end{figure*}

In this section, we first review the preliminaries of the Optimal Transport problem in Sec.~\ref{sec:optimal_transport}. Then we demonstrate how to first establish the token anchors in Sec.~\ref{sec:token_anchors} and formulate the token sets optimization strategy within intra- and inter-frame level to exploit the necessary local and global spatiotemporal contexts in Sec.~\ref{sec:spatiotemporal}, leading to compact yet effective video tokens reduction in a training-free manner. We provide an overview of our AOT in Fig.~\ref{fig:method}.

\subsection{Optimal Transport}
\label{sec:optimal_transport}

Optimal transport~(OT) distance is a widely used metric for the comparison of distributions. Here, we only focus on the discrete situation which is more related to our pipeline. Assuming we have two sets of tokens (features), the discrete distributions are formulated as:

\begin{equation} 
\label{eq:discrete}
    U=\sum_{m=1}^{M}u_m\delta_{\bm{X}_m} \hspace{2em} \text{and} \hspace{2em} V=\sum_{n=1}^{N}v_n\delta_{\bm{X}_n},
\end{equation}
where $\bm{u}$ and $\bm{v}$ are the discrete probability vectors that sum to 1, and $\delta_{\bm{X}}$ is a Dirac delta function placed at support point $\bm{X}=\{X_1,...,X_N\} \in \mathbb{R}^{N \times d}$ in the visual token embedding space.
Then, the total distance is modeled as:
\begin{equation} 
\label{eq:cost}
    <\bm{T},\bm{C}> = \sum_{m=1}^{M}\sum_{n=1}^{N}\bm{T}_{m,n}\bm{C}_{m,n},
\end{equation}
We call $\bm{C}$ the cost matrix in which each point denotes the cost between $\bm{X}_m$ and $\bm{X}_n$, such as $\bm{C}_{m,n}= 1- \text{sim}(\bm{X}_m,\bm{X}_{n})$. While $\bm{T} $ is denoted as transport plan, which is learned to minimize the total distance. The optimization problem of optimal transport is formulated as:
\begin{equation} 
\label{eq:optimization}
    \begin{aligned}
        & d_{\text{OT}}(\bm{u},\bm{v}|\bm{C}) = \underset{\bm{T}}{\text{minimize}}
         <\bm{T},\bm{C}>, \\
        & \text{subject to}
        ~~~~~~\bm{T}\bm{1}_N = \bm{u},\; \bm{T}^\top \bm{1}_M = \bm{v},\; \bm{T} \in \mathbb{R}^{M \times N}_+.
    \end{aligned}
\end{equation}

As directly optimizing the above objective requires significant time-consumption, we apply the Sinkhorn distance~\citep{cuturi2013sinkhorn}, a fast iterative solution, to use an entropic constraint for fast optimization. The optimization problem with a Lagrange multiplier of the entropy
constraint is:
\begin{equation} 
\label{eq:Sinkhorn}
    \begin{aligned}
        & d_{\text{OT},\lambda}(\bm{u},\bm{v}|\bm{C})=\underset{\bm{T}}{\text{minimize}}
         <\bm{T},\bm{C}> - \lambda h(\bm{T}), \\
        & \text{subject to}
        ~~~~~~~\bm{T}\bm{1}_N = \bm{u},\; \bm{T}^\top \bm{1}_M = \bm{v},\;
        \bm{T} \in \mathbb{R}^{M \times N}_+.
    \end{aligned}
\end{equation}
where $h(\cdot) $ is entropy and $\lambda \geq 0$ is a hyper-parameter. Then we can have a fast and off-the-shelf optimization solution with a few iterations as:
\begin{equation} 
\label{eq:Sinkhorn_optimization}
    \bm{T}^*= \text{diag}(\bm{u}^{(t)}) \exp(-\bm{C}/\lambda) \text{diag}(\bm{v}^{(t)}),
\end{equation}
where $t$ denotes the iteration and in each iteration 
$\bm{u}^{(t)} =\bm{u}/\left((\exp(-\bm{C}/\lambda)\bm{v}^{(t-1)}\right)$ and  $\bm{v}^{(t)} =\bm{v}/\left((\exp(-\bm{C}/\lambda)^\top\bm{u}^{(t)}\right) $, with the initiation $\bm{v}^{(0)} = \bm{1}$.

\subsection{Local-Global Token Anchors Establishment}
\label{sec:token_anchors}

\noindent \textbf{Global Anchors.}
Given that the final layers of visual encoders capture global information, we follow recent works~\cite{zhang2024cls,yang2025visionzip} to select global tokens that receive the most attention from the [\texttt{CLS}] token $x_{[\mathtt{CLS}]}$ in the output layer.
Specifically, the [\texttt{CLS}] attention computation for each attention head can be represented by:
\begin{equation}
\label{eq:cls-attn}
    \begin{aligned}
        S_{[\mathtt{CLS}]}^h &= \mathtt{Softmax}\!\left(\frac{Q_{[\mathtt{CLS}]}K_V^\top}{\sqrt{D}} \right),
    \end{aligned}
\end{equation}
where $Q_{[\mathtt{CLS}]}$ and $K_V$ represent the query and key output for head $h\in [1, H]$, $D$ denotes the hidden state size, and $S_{[\mathtt{CLS}]}^h$ represents the [\texttt{CLS}] attention. Given the visual token sequence $X_V$, the global tokens are then selected by:
\begin{equation}
\begin{split}
    S_{[\mathtt{CLS}]}^{\texttt{avg}}
    &= \frac{1}{H} \sum_{h=1}^H S_{[\mathtt{CLS}]}^h,\\
    \mathbf{x}_{V}^{\mathtt{g}}
    &= \operatorname{TopK}\big(\mathbf{x}_V,\; S_{[\mathtt{CLS}]}^{\texttt{avg}},\; K\big),
\end{split}
\end{equation}
where $\operatorname{TopK}(\cdot)$ selects the $K$ visual tokens with the highest $S_{[\mathtt{CLS}]}^{\texttt{avg}}$ scores, yielding the final kept token set $\mathbf{x}_{V}^{\mathtt{g}}$ of size $K$. For LVLMs without a [\texttt{CLS}] token (\textit{e.g.,} SigLip~\cite{zhai2023sigmoid} and Qwen-2.5-VL~\cite{bai2025qwen2}), we similarly define token importance scores based on self-attention (\textit{i.e.,} the average attention each
visual token receives from others) and apply the same Top-$K$ selection.

\noindent \textbf{Local Anchors.}
To enable fine-grained local details preservation, we divide the image feature into $W$ non-overlapping grid-wise windows and select locally important tokens with the highest [\texttt{CLS}] attention from a shallow layer $l$ within each window, following~\cite{li2024llava,chen2024far,zhang2024llavanextvideo}.
Given a total local budget $K$, we allocate $K_w=K/W$
tokens per window and also perform Top-$K$ selection:
\begin{equation}
    \mathbf{x}_V^{\mathtt{l}}
    =
    \bigcup_{w=1}^W
    \operatorname{TopK}\big(
        \mathbf{x}_V^w,\,
        S_{[\mathtt{CLS}]}^{\texttt{avg}},\,
        K_w
    \big),
\end{equation}
where $\mathbf{x}_V^w$ denotes tokens in window $w$.
The final anchor set is the union of global and local tokens,
$\mathbf{X}_V^{\texttt{anchors}} = \mathbf{x}_V^{\mathtt{g}} \cup \mathbf{x}_V^{\mathtt{l}}$,
and the remaining tokens are denoted as
$\mathbf{X}_V^{\texttt{unanchors}}$.
Following~\cite{zhang2025vscan}, we balance global and local selection, \textit{e.g.,} $|\mathbf{x}_V^{\mathtt{g}}| = |\mathbf{x}_V^{\mathtt{l}}|$, and exclude locally selected tokens from the global set to avoid duplication and redundancy.

\subsection{Spatiotemporal Pruning}
\label{sec:spatiotemporal}

\noindent \textbf{Intra-Frame Pruning with OT.}
For each frame, given extracted visual tokens as $\bm{X}=\{X_1,...,X_N\} \in \mathbb{R}^{N \times d}$, we perform token anchors selection strategy above to establish $\mathbf{X}_V^{\texttt{acnhors}} \in \mathbb{R}^{M \times d}$ (dubbed $\mathbf{X}_V^{\texttt{a}}$) and $\mathbf{X}_V^{\texttt{unanchors}} \in \mathbb{R}^{(N-M) \times d}$ (dubbed $\mathbf{X}_V^{\texttt{u}}$). Built upon OT, we learn the geometric alignment transport plan $\bm{T}$ with these fixed support sets $\mathbf{X}_V^{\texttt{a}}$ and $\mathbf{X}_V^{\texttt{u}}$, by minimizing the following OT distance to push $\mathbf{X}_V^{\texttt{u}}$ to $\mathbf{X}_V^{\texttt{a}}$:
\begin{equation}
\label{eq:distance}
    \begin{aligned}
         d^{intra}_{\text{OT}}(k) = d_{\text{OT}}(\bm{u},\bm{v}|\bm{1}-\bm{{(X^a_V)}}^\top\bm{({X}^{u}_{V})}),
    \end{aligned}
\end{equation}
where $\bm{C}=\bm{1}-\bm{{(X}^a_V)}^\top\bm{({X}^{u}_{V})}$ denotes that we use the inverse cosine similarity distance between $\bm{X^a_V}$ and $\bm{X^u_V}$ as the cost matrix. Then we compute the solution of transport plan $\bm{T_{intra}}^*$ as Eq.~\ref{eq:Sinkhorn_optimization} and the final OT distance $d^{intra}_{\text{OT}}(k)$. Given the optimal transport plan $\bm{T_{intra}}^* \in \mathbb{R}_+^{(N-M)\times M}$ between
$\mathbf{X}_V^{\texttt{u}}$ and $\mathbf{X}_V^{\texttt{a}}$, we aggregate the unselected tokens onto the anchor tokens to obtain the compressed visual representation.
For the $j$-th anchor token, we first compute its received transport mass from all unselected tokens:
\vspace{-1mm}
\begin{equation}
    m_j \;=\; \sum_{i=1}^{N-M} T^*_{ij},
\vspace{-1mm}
\end{equation}
where mass denotes the supplied context units and then update the corresponding anchor by a mass-normalized OT aggregation from unselected tokens:
\begin{equation}
    \tilde{\mathbf{x}}^{a}_j
    \;=\;
    \frac{
        \mathbf{x}^{a}_j
        \;+\;
        \lambda_{intra} \sum_{i=1}^{N-M} T^*_{ij} \,\mathbf{x}^{u}_i
    }{
        1 \;+\; \lambda_{intra} m_j
    },
    \quad j = 1,\dots,M,
\end{equation}
where $\lambda_{intra}$ is a weighting coefficient for controlling the final contextual contribution to token anchors of the OT-based update. The final intra-frame compressed token set is then updated as:
\begin{equation}
    \tilde{\mathbf{X}}_V^{\texttt{a}}
    \;=\;
    \left\{\tilde{\mathbf{x}}^{a}_1, \dots,
    \tilde{\mathbf{x}}^{a}_M\right\}
    \in \mathbb{R}^{M \times d},
\end{equation}
which is used as the pruned visual tokens for the subsequent temporal pruning.

\noindent \textbf{Inter-Frame Pruning with OT.}
Beginning by segmenting the overall sampled frames into several frame clips which is adopted by uniform sampling or dynamic clustering, for each frame clip $\mathcal{C} = \{t_1,\dots,t_L\}$, we use the intra-frame compressed tokens of the first frame as temporal anchors:
\[
\mathbf{A}^{(1)} = \tilde{\mathbf{X}}^{\texttt{a}}_{V,t_1}
= \{\mathbf{x}^{(1)}_j\}_{j=1}^M.
\]
For each subsequent frame $t_\ell$ ($\ell = 2,\dots,L$), let
$\mathbf{S}^{(\ell)} = \tilde{\mathbf{X}}^{\texttt{a}}_{V,t_\ell}
= \{\mathbf{x}^{(\ell)}_i\}_{i=1}^{M}$ denotes its intra-frame
compressed tokens, and $\mathbf{A}^{(\ell-1)} = \{\mathbf{x}^{(\ell-1)}_j\}_{j=1}^{M}$ represents the current clip anchors after $(\ell-1)$ times OT aggregation across consecutive frames.
For each subsequent frame $t_\ell$ ($\ell=2,\dots,L$), with anchors
$\mathbf{A}^{(\ell-1)} = \{\mathbf{x}^{(\ell-1)}_j\}_{j=1}^{M}$ and
tokens $\mathbf{S}^{(\ell)} = \{\mathbf{x}^{(\ell)}_i\}_{i=1}^{M}$,
we further formulate the inter-frame OT distance using optimal transport as:
\begin{equation}
    d^{\text{inter}}_{\text{OT}}(\ell)
    =
    d_{\text{OT}}\!\left(
        \bm{u}, \bm{v}
        \,\big|\,
        \bm{1}
        -
        \big(\mathbf{A}^{(\ell-1)}\big)
        \big(\mathbf{S}^{(\ell)}\big)^\top
    \right),
\end{equation}
and obtain the inter-frame optimal transport plan as:
\begin{equation}
    \bm{T_{inter}}^{(\ell)\,*}
    =
    \mathrm{OT}\!\left(
        \bm{u}, \bm{v}
        \,\big|\,
        \bm{1}
        -
        \big(\mathbf{A}^{(\ell-1)}\big)
        \big(\mathbf{S}^{(\ell)}\big)^\top
    \right),
\end{equation}
where this also utilizes a fast Sinkhorn-based solver similar to the intra-frame scenario. We then apply normalization along the row of $\bm{T_{inter}}^{(\ell)\,*}$ to get the assignment probabilities,
\begin{equation}
    p^{(\ell)}_{ij}
    =
    \frac{T^{(\ell)\,*}_{ij}}{\sum_{j'} T^{(\ell)\,*}_{ij'}},
    \qquad
    q^{(\ell)}_i = \max_j p^{(\ell)}_{ij},
\end{equation}
which are used to decide whether a token is aggregated into anchors
who showcases potential similar visual content or kept as a temporally variant token to maintain dynamics.
Specifically, if $q^{(\ell)}_i < \tau$, the $i$-th token of frame $t_\ell$
is considered having drastic temporal change and kept as usual; otherwise, it is smoothly aggregated into the current clip anchors analogous to intra-frame pruning optimization:
\begin{equation}
    \mathbf{a}^{(\ell)}_j
    =
    \frac{
        \mathbf{a}^{(\ell-1)}_j
        +
        \lambda_{inter}
        \sum_{i:\,q^{(\ell)}_i \ge \tau}
        p^{(\ell)}_{ij}\,\mathbf{s}^{(\ell)}_i
    }{
        1
        +
        \lambda_{inter}
        \sum_{i:\,q^{(\ell)}_i \ge \tau}
        p^{(\ell)}_{ij}
    },
    \quad j=1,\dots,M.
\end{equation}
By iterating this procedure over frames in the clip, we obtain
one compact set of clip-level anchors (from $\mathbf{A}^{(\ell)}$) along with a small number of remaining high-change tokens, which together facilitate the spatiotemporal pruned visual tokens.
The proposed AOT obtains necessary local-global semantic and context aggregation across spatiotemporal dimensions by adopting optimal transport to assemble informative cues from low-discriminative or relatively similar tokens under removing or merging. Our method considers their intrinsic contributions to the compact token anchors which globally abstracts and compresses spatiotemporal redundancy and retains essential temporal dynamics, significantly accelerating Video LLM inference while preserving both temporal and visual integrity. Meanwhile, optimal transport solution can be quickly and efficiently solved by the off-the-shelf Sinkhorn-Knopp Iteration~\cite{cuturi2013sinkhorn} with negligible computation overhead, leading to an efficient and effective training-free manner.

\section{Experiments}
\label{sec:experiments}

\subsection{Experimental Settings} \label{sec:experimental_settings}
\noindent \textbf{Benchmarks.}
To demonstrate the effectiveness and efficiency of our proposed AOT, we evaluate our method on four commonly used video understanding benchmarks: MVBench~\cite{li2024mvbench}, EgoSchema~\cite{mangalam2023egoschema}, LongVideoBench~\cite{wu2024longvideobench}, and VideoMME~\cite{fu2025video}. Comprising videos of various lengths and complex scenarios, these benchmarks provide a comprehensive testbed for demonstrating the effectiveness and generalization of our method.

\begin{table*}[t]
\centering
\caption{ 
Comparison of state-of-the-art methods on LLaVA-OneVision~\cite{li2024llava} across video benchmarks. The best performance among those with similar retention ratios is highlighted in \textbf{bold}, while the second best will be denoted as \underline{underlined}.
}
\vspace{-0.2cm}
\resizebox{\textwidth}{!}{%
\begin{tabular}{l|ccc|cccc|cc}
\toprule
\multirow{2}{*}{Method} &
\multirow{2}{*}{\makecell{Prefilling\\FLOPs (T) $\downarrow$}} & \multirow{2}{*}{\makecell{FLOPs\\Ratio $\downarrow$}} & \multirow{2}{*}{\makecell{Before LLM\\Retained Ratio}}
& \multirow{2}{*}{\makecell{MVBench\\ $\uparrow$}} & \multirow{2}{*}{\makecell{EgoSchema\\ $\uparrow$}} & \multirow{2}{*}{\makecell{LongVideo\\Bench $\uparrow$}} & \multirow{2}{*}{\makecell{VideoMME\\ $\uparrow$}} & \multicolumn{2}{c}{Avg. $\uparrow$}  \\
&&&&&&&& Score & \% \\
\midrule
\rowcolor{gray!20} 
LLaVA-OV-7B & 40.8 & 100\% & 100\% 
& 58.3 & 60.4 & 56.4 & 58.6 & 58.4 & 100 \\ %
FastV~\cite{chen2024image} & 9.3 & 22.8\% & 100\%
& 55.9 & 57.5 & \textbf{56.7} & 56.1 & 56.5 & 96.7 \\ %
PDrop~\cite{xing2024pyramiddrop} & 10.5 & 25.7\% & 100\%
& 56.1 & 58.0 & 54.1 & 56.4 & 56.2 & 96.2 \\
DyCoke \cite{tao2025dycoke} & 8.7 & 21.3\% & 25\% & 53.1 & 59.5 & 49.5 & 54.3 & 54.1 & 92.6 \\

VisionZip~\cite{yang2025visionzip} & 8.7 & 21.3\% & 25\%
& \underline{57.9} & \underline{60.3} & \underline{56.5} & \textbf{58.2} & \underline{58.2} & \underline{99.7} \\ %

PruneVid~\cite{huang2024prunevid} & 8.7 & 21.3\% & 25\%
& 57.4 & 59.9 & 55.7 & 57.4 & 57.6 & 98.6 \\ %

FastVID~\cite{shen2025fastvid} & 8.7 & 21.3\% & 25\%
& 56.5 & - & 56.3 & \underline{58.0} & - & - \\

\textbf{AOT} & 8.7 & 21.3\% & 25\%
& \textbf{58.7} & \textbf{61.3} & 56.3 & 57.5 & \textbf{58.5} &  \textbf{100.0} \\ %
\midrule
VisionZip~\cite{yang2025visionzip} & 7.0 & 17.2\% & 20\%
& \underline{57.7} & \underline{59.8} & 55.2 & \textbf{57.9} & \underline{57.7} & \underline{98.8} \\ %

PruneVid~\cite{huang2024prunevid} & 7.0 & 17.2\% & 20\%
& 57.2 & 59.7 & 54.7 & 56.9 & 57.1 & 97.8 \\ %

FastVID~\cite{shen2025fastvid} & 7.0 & 17.2\% & 20\%
& 56.3 & - & \textbf{57.1} & \textbf{57.9} & - & - \\

\textbf{AOT} & 7.0 & 17.2\% & 20\%
& \textbf{58.1}  & \textbf{61.3} & \underline{56.2} & \underline{57.2} & \textbf{58.2} & \textbf{99.7} \\ %
\midrule
VisionZip~\cite{yang2025visionzip} & 5.2 & 12.7\% & 15\%
& 56.5 & \underline{59.8} & 54.4 & 56.1 & 56.7 & 97.1 \\ %

PruneVid~\cite{huang2024prunevid} & 5.2 & 12.7\% & 15\%
& \underline{56.8} & 59.7 & \underline{55.4} & \underline{56.6} & \underline{57.1} & \underline{97.8} \\ %

FastVID~\cite{shen2025fastvid} & 5.2 & 12.7\% & 15\% & 56.0 & - & \textbf{56.2} & \textbf{57.7} & - & - \\

\textbf{AOT} & 3.4 & 8.3\% & 15\%
& \textbf{57.8}  & \textbf{61.3} & 55.2 & \underline{56.6} & \textbf{57.7} & \textbf{98.8} \\ %
\midrule
VisionZip~\cite{yang2025visionzip} & 3.4 & 8.3\% & 10\% 
& 53.5 & 58.0 & 49.3 & 53.4 & 53.5 & 91.6 \\ %

PruneVid~\cite{huang2024prunevid} & 3.4 & 8.3\% & 10\% 
 & \underline{56.2} & \underline{59.8} & \underline{54.5} & 56.0 & \underline{56.6} & \underline{96.9} \\ %

FastVID~\cite{shen2025fastvid} & 3.4 & 8.3\% & 10\% 
& 55.9 & - & \textbf{56.3} & \textbf{57.3} & - & - \\

\textbf{AOT} & 5.2 & 12.7\% & 10\%
& \textbf{57.0}  & \textbf{60.6} & 54.2 & \underline{56.1} & \textbf{57.0} & \textbf{97.6} \\ %
\bottomrule

\end{tabular}
}
\label{tab:ov_benchmark}
\vspace{-0.2cm}
\end{table*}

\begin{table*}[t]
\centering
\caption{Comparison of state-of-the-art methods on LLaVA-Video~\cite{zhang2024video} across video benchmarks. The best performance among those is highlighted in \textbf{bold}, while the second best will be denoted as \underline{underlined}, demonstrating consistent effectiveness.
}
\vspace{-0.2cm}
\resizebox{\textwidth}{!}{%
\begin{tabular}{l|ccc|cccc|cc}
\toprule
\multirow{2}{*}{Method} &
\multirow{2}{*}{\makecell{Prefilling\\FLOPs (T) $\downarrow$}} & \multirow{2}{*}{\makecell{FLOPs\\Ratio $\downarrow$}} & \multirow{2}{*}{\makecell{Before LLM\\Retained Ratio}}
& \multirow{2}{*}{\makecell{MVBench\\ $\uparrow$}} & \multirow{2}{*}{\makecell{EgoSchema\\ $\uparrow$}} & \multirow{2}{*}{\makecell{LongVideo\\Bench $\uparrow$}} & \multirow{2}{*}{\makecell{VideoMME\\ $\uparrow$}} & \multicolumn{2}{c}{Avg. $\uparrow$}  \\
&&&&&&&& Score & \% \\
\midrule
 \cellcolor{gray!20}{LLaVA-Video-7B} & \cellcolor{gray!20}{80.2} & \cellcolor{gray!20}{100\%} & \cellcolor{gray!20}{100\%} 
 & \cellcolor{gray!20}{60.4} & \cellcolor{gray!20}{57.2} & \cellcolor{gray!20}{58.9} & \cellcolor{gray!20}{64.3} & \cellcolor{gray!20}{60.2} & \cellcolor{gray!20}{100} \\ %

 FastV~\cite{chen2024image} & 17.1 & 21.3\% & 100\% 
& 54.3 & 54.1 & \underline{55.0} & 58.8 & 55.6 & 92.4 \\ %

 PDrop~\cite{xing2024pyramiddrop} & 19.5 & 24.3\% & 100\% 
& 55.9 & 54.3 & 54.7 & \underline{61.9} & \underline{56.7} & \underline{94.2} \\ %

 VisionZip~\cite{yang2025visionzip} & 9.3 & 18.9\% & 25\%
& \underline{56.7} & \underline{54.7} & 54.7 & 60.7 & \underline{56.7} & \underline{94.2} \\ %

 DyCoke~\cite{yang2025visionzip} & 9.3 & 18.9\% & 25\%
& 50.8 & - & 53.0 & 56.9 & - & - \\ %

 \textbf{AOT} & 9.3 & 18.9\% & 25\%
& \textbf{58.8}  & \textbf{55.4} & \textbf{56.2} & \textbf{62.4} & \textbf{58.2} & \textbf{96.7} \\ %

\hline

 VisionZip~\cite{yang2025visionzip} & 9.3 & 11.6\% & 15\%
& \underline{56.7} & \underline{54.7} & \underline{54.7} & \underline{60.7} & \underline{56.7} & \underline{94.2} \\ %

 \textbf{AOT} & 9.3 & 11.6\% & 15\%
& \textbf{57.8}  & \textbf{55.2} & \textbf{55.0} & \textbf{62.0} & \textbf{57.5} & \textbf{95.5} \\ %
\bottomrule

\end{tabular}
}%
\label{tab:diff_backbone}
\vspace{-0.4cm}
\end{table*}

\noindent \textbf{Implementation Details.}
\label{sec:implementation_details}
Our method is implemented on LLaVA-OneVision-7B~\cite{li2024llava} and LLaVA-Video-7B~\cite{zhang2024video} models. Both evaluation and inference use NVIDIA A100 GPUs. Inference cost is measured by prefilling FLOPs, with baselines configured for comparable FLOPs (see in the appendix). The number of intra-frame token anchors is set to be 126 for a 10\% token retention budget. Both intra- and inter-frame Sinkhorn-Knopp iterations are conducted with 100. The weighting coefficient $\lambda_{intra}$ and $\lambda_{inter}$ are set to default 1.0, otherwise stated.
Following official practice, LLaVA-OneVision models utilize 32 input video frames ($N_v$ = 196), while LLaVA-Video uses 64 sampled frames ($N_v$ = 169). All benchmarks are conducted using LMMs Eval~\cite{zhang2025lmms,li2024lmms} to report the results.

\noindent \textbf{Compared Baselines.}
We evaluate our proposed \textbf{\textit{AOT}} against six representative training-free approaches.
(1) FastV~\cite{chen2024image}, which selects salient visual tokens during prefilling based on attention guidance between predicted tokens and vision tokens;
(2) PDrop~\cite{xing2024pyramiddrop}, which discards visual tokens across hierarchical LLM blocks under the guidance of image and instruction tokens;
(3) VisionZip~\cite{yang2025visionzip}, which performs spatial token merging before feeding visual tokens into the LLM;
(4) DyCoke~\cite{tao2025dycoke}, which integrates temporal token merging before LLM with adaptive KV cache pruning during decoding;
(5) PruneVid~\cite{huang2024prunevid}, which reduces redundancy via jointly spatial and temporal token clustering;
and (6) FastVID~\cite{shen2025fastvid}, a recent method that segments videos into clips and applies density-aware token pruning.

\noindent \textbf{Inference Cost Analysis.}
\label{sec:inference_cost}
In this section, we investigate the FLOPs of the LLM backbone by counting the cost of each Transformer layer (MHA+FFN). Following~\cite{chen2024image,xing2024pyramiddrop,tao2025dycoke,shao2025holitom}, processing $n_i$ vision tokens at layer $i$ with hidden size $d$ and FFN width $m$ costs $4n_i d^2 + 2n_i^2 d + 2n_i d m$ FLOPs.
For an LLM with $T$ layers, the prefilling and decoding FLOPs can be computed as:
\begin{align}
    F_{\text{pre}}
    &= \sum_{i=1}^T \big(4n_i d^2 + 2n_i^2 d + 2n_i d m\big),\\
    F_{\text{dec}}
    &= \sum_{i=1}^T
    R\Big((4d^2 + 2d m) + 2\big(d n_i + \tfrac{1}{2}d(R+1)\big)\Big),
\end{align}
and the total cost is $F_{\text{pre}} + F_{\text{dec}}$.
With token generation number $R=100$, the decoding term contributes only about \textbf{2\%} of the total FLOPs in video LLMs, showing that computation is dominated by the prefilling stage. Therefore, reducing visual tokens \textit{before} feeding them into the LLM strikes significantly larger savings than pruning applied only inside the early layers of the LLM.

\subsection{Main Results}
\label{sec:main_results}

\noindent \textbf{Results on LLaVA-OneVision.} 
As shown in the Table.~\ref{tab:ov_benchmark}, we conduct AOT evaluations on various video benchmarks against state-of-the-art approaches on the LLaVA-OneVision 7B model, comparing the performances and analyzing the inference cost (so-called FLOPs) under various token retention budgets (25\%, 20\%, 15\%, and 10\%) before feeding the visual tokens into LLM for subsequent processing. Inner-LLM pruning methods, such as FastV~\cite{chen2024image} and PDrop~\cite{xing2024pyramiddrop}, often struggle to balance performance and efficiency, especially at lower token retention ratios (25\%). DyCoke~\cite{tao2025dycoke}, which segments video frames into groups of 4 and prunes all but the first frame, is limited by its design, capping its lowest retention ratio at 25\%. Spatial pruning methods like VisionZip~\cite{yang2025visionzip} show a significant performance drop (up to 8.4\%) at 10\% retention. This decline can be attributed to relying solely on spatial compression, which is less effective at preserving crucial spatiotemporal information needed for performance under aggressive pruning. Notably, even after
pruning \textbf{90.0\%} of the visual tokens, our AOT preserves averaged \textbf{97.6\%} of the vanilla model’s performance.
This demonstrates the superior robustness and adaptability of our approach compared to existing methods.

\noindent \textbf{Results on LLaVA-Video.} Table.~\ref{tab:diff_backbone} presents our AOT's performance on LLaVA-Video 7B model. 
LLaVA-Video-7B poses a greater compression challenge due to its higher initial pooling rate (169~\textit{vs.}~196 tokens for each frame in LLaVA-OneVision). Despite this, our method achieves a reduction to just 15\% of the original FLOPs, retaining 95.5\% performance and outperforming existing methods. 
Generally speaking, obtaining significant token compression with minimal performance decrease is indeed more difficult for LLaVA-Video 7B than for LLaVA-OneVision 7B.

\noindent \textbf{Scaling Input with more frames.}
To validate the performance fluctuation when varying input frames, we conduct experiments with increasing input frames to present the robust improvements of our method in Fig.~\ref{fig:scaling_frames}. A crucial challenge for video LLMs is that uniformly and sparsely sampled frames may lose essential visual information to obtain accurate answers. 
Fig.~\ref{fig:scaling_frames} illustrates that our approach consistently surpasses other compression methods across frame rates. 
At the sparse 16 sampled frames that contain less temporal redundancy, our AOT still transcends all other compression approaches. With denser 64 frames, thanks to our intra- and inter-level optimal token pruning, our method showcases more efficient and effective performances over the vanilla model. Moreover, when inputting with 128 frames, our token reduction method maintains an approximate context length, whereas the vanilla models suffer from maximum context length limitations. This further demonstrates the superiority of our proposed AOT that excels at tasks requiring extensive spatiotemporal context or answering complex questions with long text.

\begin{figure}[t]
    \centering
    \includegraphics[width=\linewidth]{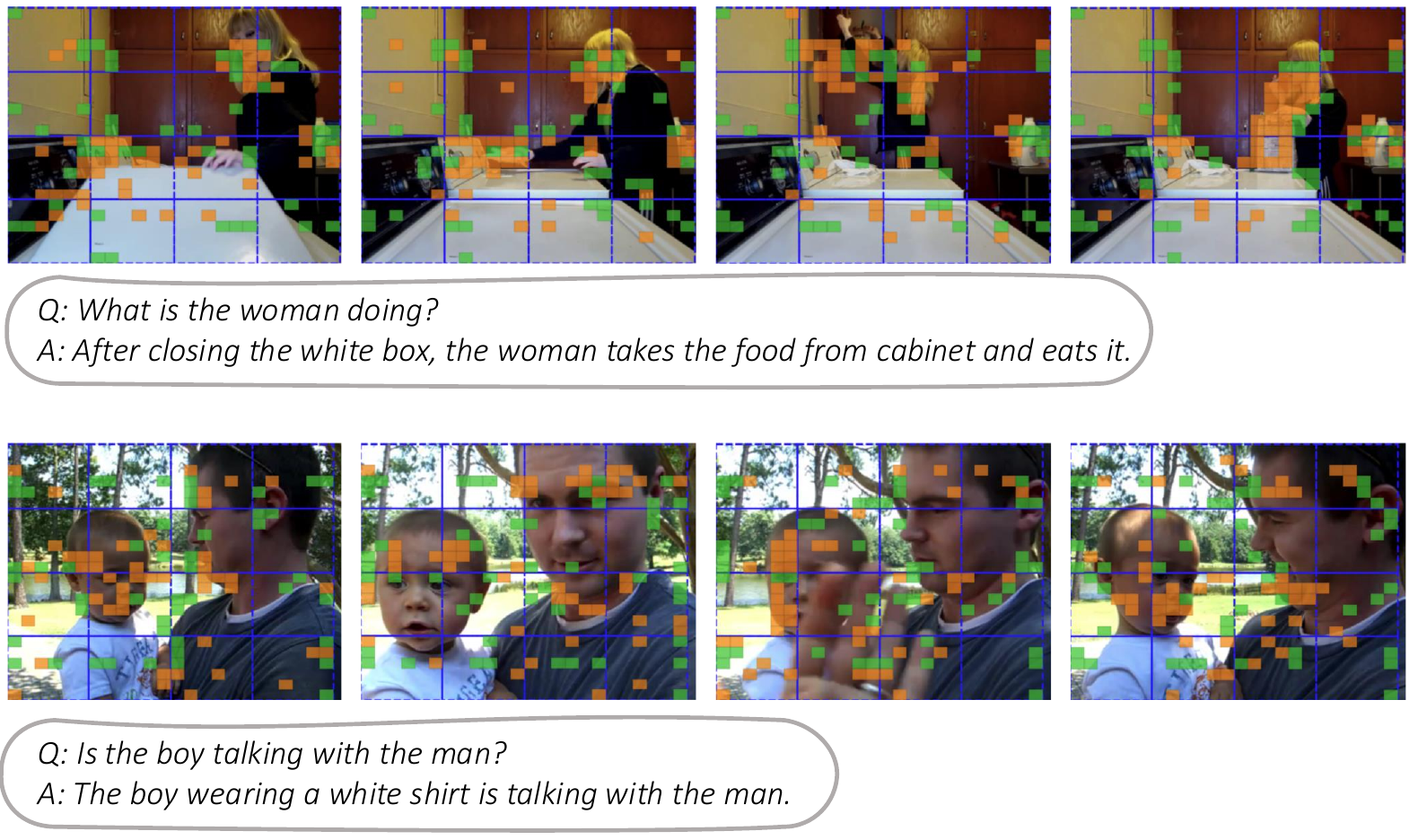}
    \vspace{-0.4cm}
    \caption{Qualitative visualizations of our \textcolor{green}{Local}-\textcolor{orange}{Global} token anchors evolution across consecutive frames while optimal transport is adopted to aggregate necessary information from unselected tokens to help LLM precess better.}
    \label{fig:vis_cases}
    \vspace{-0.4cm}
\end{figure}

\noindent \textbf{Superior Performances after Token Reduction.}
As shown in Table.~\ref{tab:ov_benchmark} and Fig.~\ref{fig:scaling_frames}, we can find that models utilizing our token reduction can surpass the vanilla models on some benchmarks. This elucidates that there is excessive, irrelevant, or redundant information acting as noise, inducing useless visual cues for effective and efficient abstraction. The massive visual information misguides the subsequent LLM to focus on the critical parts, hence resulting in degraded results. Rather than directly removing unimportant or merging highly similar tokens, our method exploits optimal transport to comprehensively mine out informative signals from these tokens that consist of essential semantic and context to boost the selected token anchors, via a transport-weighted aggregation. In this way, the results highlight the efficacy of our method in aggregating or \textit{distilling} key visual information and further globally optimal refinement for better LLM processing, while we present qualitative visualizations in Fig.~\ref{fig:vis_cases}.

\subsection{Efficiency Analysis Sinkhorn-Knopp Iteration.} 
As shown in Table.~\ref{tab:ota_time}, we measure time consumption under 32 frames scenario using LLaVA-OneVision 7B on one single A100 GPU in terms of optimal transport using Sinkhorn-Knopp Iteration. We can see that with 100 iterations, the computational time for Intra-Frame OT takes around 0.51 milliseconds while 1.60 milliseconds to finish Inter-Frame OT, summing up to nearly 2.11 milliseconds which takes less than \textbf{1\%} of total inference time. For VisionZip~\cite{yang2025visionzip}, PruneVid~\cite{huang2024prunevid}, and our proposed AOT require token pre-processing before LLM, which introduces extra time. However, our method constructs a sparse and fixed number of tokens per frame as token anchors, while PruneVid produces a variable number of tokens per frame that complicates the batch processing acceleration, inevitably inducing extra computation overhead.

\subsection{Ablation Studies}
\label{sec:ablatives}

In this section, we conduct ablation studies on LLaVA-OneVision 7B by setting the token retention budget at 10\% to gradually demonstrate the improvements from each of our proposed components.

\noindent \textbf{Ablation on Intra-Frame Token Anchors Selection.}
To validate the effectiveness of the Local-Global token anchors establishment, shown in Table.~\ref{tab:ablations}, despite leveraging global-level anchors only can reach competitive performances across video benchmarks after employing our AOT, combining both global and local anchors achieves the best, demonstrating the effectiveness of retaining semantically important and spatially diverse token candidates.

\begin{figure}[t]
    \centering
    \includegraphics[width=1.0\linewidth]{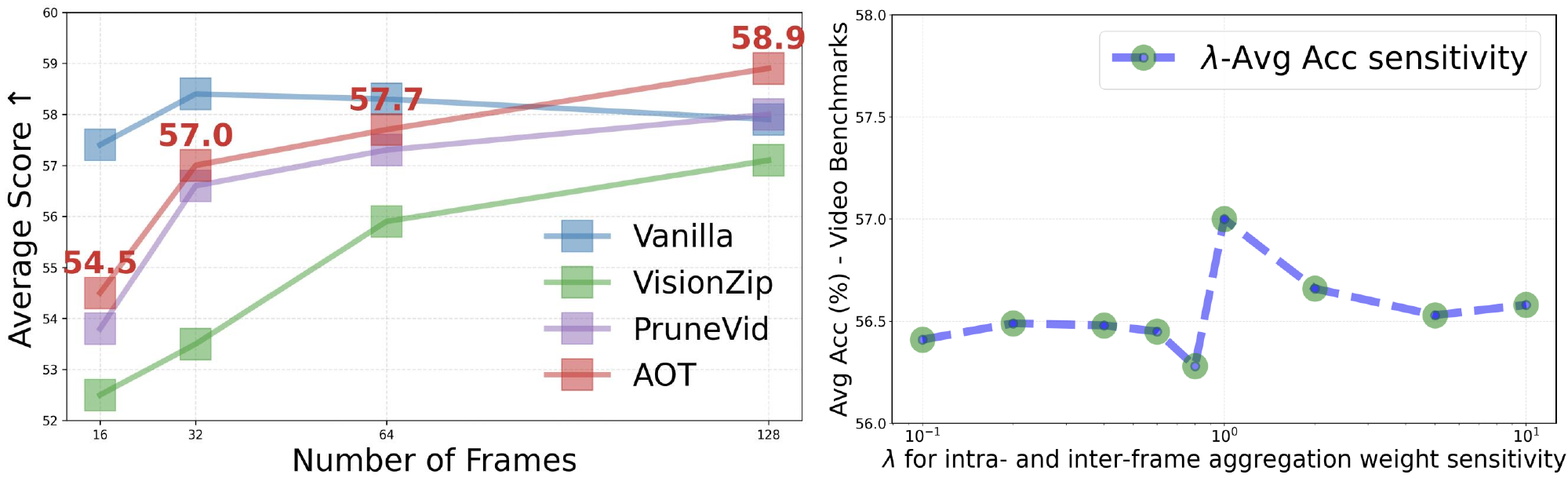}
    \vspace{-0.5cm}
    \caption{\textbf{Left:} scaling with more frames leads to more efficient and effective visual information abstraction. \textbf{Right:} sensitivity analysis of weighting coefficient controlling contextual contribution with consistent configuration, $\lambda_{intra}$ and $\lambda_{inter}$.}
    \label{fig:scaling_frames}
    \vspace{-0.4cm}
\end{figure}

\noindent \textbf{Ablation on OT for Token Reduction.} In Table.~\ref{tab:ablations}, we can also observe that without optimal transport to aggregate comprehensive semantics and context from pruned tokens, the model will result in degraded performances, particularly for aggressive token compression, elucidating the importance of these subtle yet informative context.

\noindent \textbf{Ablation on OT for Intra-Frame and Inter-Frame.}
We conduct experiments in Table.~\ref{tab:ablations} that integrating the optimal transport from both intra- and inter-frame optimization helps aggregate necessary semantics and contexts from merging or removing tokens into selected token anchors to process complex videos better, instead of applying simple merging or removing them. This further demonstrates our approach significantly accelerates Video LLM inference while preserving both temporal and visual integrity.

\noindent \textbf{Ablation on Aggregation for Token Anchors.}
In Table.~\ref{tab:merging_strategy}, we further conduct studies to validate the effectiveness of aggregating the information from unimportant and very similar tokens through the optimized transport plan $T$, including intra- and inter-frame. Compared with no merging and cosine similarity weighting, the transported-weighting $T$ showcases superior token compression performances, which demonstrates the efficacy of local-global Optimal Transport aggregation in this paper.

\noindent \textbf{Sensitivity on Weightings $\lambda_{Intra}$ and $\lambda_{Inter}$.} As shown at the right of Fig.~\ref{fig:scaling_frames}, both $\lambda_{Intra}$ and $\lambda_{Inter}$ with default 1.0 strikes the best, while much lower weightings aggregate marginal information and higher ones induce noise, both of them result in degraded performances.

\begin{table}[tb]
    \centering
    \caption{Computational overhead in terms of Intra- and Inter-Frame using Sinkhorn-Knopp Iteration (milliseconds).}
    \vspace{-0.2cm}
    \setlength\tabcolsep{2pt} 
    \resizebox{0.70\linewidth}{!}{%
    \begin{tabular}{l|ccc|c}
    \toprule
    {Iteration}& {Intra-Frame} & {Inter-Frame} & {Overall} & \% \\
    \midrule
    50      & 0.50~ms & 1.50~ms & 2.00~ms & $\le$ 1\%\\
    100     & 0.51~ms & 1.60~ms & 2.11~ms & $\le$ 1\%\\
    \bottomrule
    \end{tabular}%
    }
    \vspace{-0.2cm}
    \label{tab:ota_time}
\end{table}

\begin{table}[tb]
    \centering
    \caption{Ablation: The effect of different aggregation strategy to obtain the final token anchors set to represent the video.}
    \setlength\tabcolsep{2pt} 
    \vspace{-0.2cm}
    \resizebox{1.0\linewidth}{!}{%
    \begin{tabular}{l|cccc|c}
    \toprule
    {Aggregation}& {MVBench} & {EgoSchema} & {LongVideo Bench} & {VideoMME} & {Avg.~$\uparrow$}\\
    \midrule
    No Merging      & 56.1 & 60.2 & 53.5 & 55.8 & 56.4 \\
    Cosine Merging  & 51.5 & 55.8 & 51.1 & 51.3 & 52.4 \\
    Ours AOT        & 57.0 & 60.6 & 54.2 & 56.1 & 57.0 \\
    \bottomrule
    \end{tabular}%
    }
    \vspace{-0.2cm}
    \label{tab:merging_strategy}
\end{table}

\begin{table}[tb]
\centering
\caption{Ablation: Contribution of each component by gradually removing the proposed component to demonstrate effectiveness.}
\vspace{-0.2cm}
\resizebox{0.48\textwidth}{!}{%
\begin{tabular}{l|cccc|cc}
\toprule
\multirow{1}{*}{Method} & \multirow{1}{*}{\makecell{MVBench}} & \multirow{1}{*}{\makecell{EgoSchema}} & \multirow{1}{*}{\makecell{LongVideo\\Bench}} & \multirow{1}{*}{\makecell{VideoMME}} & \multicolumn{2}{c}{Avg.}  \\
&&&&& Score~$\uparrow$ & \%~$\uparrow$ \\
\midrule
\rowcolor{gray!20} 
Vanilla & 58.3 & 60.4 & 56.4 & 58.6 & 58.4 & 100 \\ %

\hline

\textit{w/o} Local Anchors & 56.5 & 60.1 & 54.0 & 55.7 & 56.6 & 96.9 \\ %
 \textit{w/o} Global Anchors & 55.5 & 59.4 & 53.4 & 53.1 & 55.4 & 94.9 \\ %

\hline

 \textit{w/o} OT & 56.1 & 60.2 & 53.5 & 55.8 & 56.4 & 96.6 \\ %

\hline

 OT \textit{w/o} Intra-frame & 57.1 & 60.2 & 53.6 & 54.6 & 56.3 & 96.6 \\ %
 OT \textit{w/o} Inter-frame & 56.1 & 60.0 & 53.6 & 55.9 & 56.4 & 96.6 \\ %

\hline

AOT & 57.0 & 60.6 & 54.2 & 56.1 & 57.0 & 97.6 \\ %

\bottomrule
\end{tabular}
}
\label{tab:ablations}
\vspace{-0.2cm}
\end{table}

\section{Conclusion}
\label{sec:conclusion}

In this paper, we first investigate how to aggregate necessary yet optimal semantics and contexts from merging or removing tokens into remaining tokens, instead of simply merging or removing them. To start with semantically important and spatially diverse token candidates, we perform a local-global token selection to establish token anchors for each frame. Then, we utilize an optimal transport strategy to comprehensively aggregate spatiotemporal context from the transport plan within intra- and inter-frame to preserve temporal and visual fidelity with a training-free pipeline. In this way, our method presents competitive performances across various video benchmarks under aggressive token compression, which hopes to shed new light on the community.

\clearpage
\section*{Acknowledgments}
This work was supported by EU Horizon projects ELIAS (No. 101120237) and ELLIOT (No. 101214398) by the FIS project GUIDANCE (No. FIS2023-03251).
We acknowledge ISCRA for awarding this project access to the LEONARDO supercomputer, owned by the EuroHPC Joint Undertaking, hosted by CINECA (Italy). We acknowledge EuroHPC Joint Undertaking for awarding us access to MareNostrum5 as BSC, Spain.


\begin{center}
    \section*{------ Supplementary Material ------}
\end{center}

\begin{abstract}
This supplementary material provides additional details and analysis to support the main paper, as follows:
\begin{itemize}
    \item In Sec.~\ref{sec:ot_analysis}, we provide with more detailed theoretical analysis in terms of optimal transport.
    \item In Sec.~\ref{sec:more_imples}, we describe more detailed implementations.
    \item In Sec.~\ref{sec:dynamic_clip}, we conduct more experiments on dynamic frame clip setting to demonstrate the practical advantage of our proposed AOT.
    \item In Sec.~\ref{sec:random}, we conduct more ablation study by randomly selecting the token anchors within intra-frame level to illustrate the importance of high-quality token anchors establishment.
    \item In Sec.~\ref{sec:visualization}, we provide with more visualizations in terms of our initial token anchors.
    \item In Sec.~\ref{sec:limit_future}, we support with the limitation of our method and future improvements.
\end{itemize}

\end{abstract}

\section{AOT Approach Details}
\label{sec:ot_analysis}

\subsection{Optimal Transport}

The Optimal Transport~\citep{monge1781memoire} is preliminarily introduced to find out a transportation plan that delivers several products to satisfy the demand of various consumers with a minimal cost, such as pouring beverage between several containers until all of them are filled. Recently, it has been widely utilized to make comparisons of distributions. Specifically, given two probability density function $U$ and $V$ over space $\mathcal{X} $ and $\mathcal{Y} $, then the OT (Wasserstein) distance~\citep{thorpe2018introduction} can be defined as:
\begin{equation} 
\label{eq:wasserstein}
    D_{\text{OT}}(U,V) = \underset{\Gamma}{\inf}  \int_{\mathcal{X}\times\mathcal{Y} } \bm{C}(\bm{x},\bm{y}) d\gamma(\bm{x},\bm{y}),
\end{equation}
where $\bm{C}(\bm{x},\bm{y})$ is the cost between two points in the space $\mathcal{X}\times\mathcal{Y} $, and $\Gamma $ denotes the set of transport plans between support points $ \bm{x}$ and $ \bm{y}$ (\textit{e.g.,} $\gamma(\bm{x},\bm{y}) $). 
We can regard two probability density functions $U$ and $V$ as the beverage and containers, and $\bm{C}$ is the cost function of pouring a unit of beverage.

When it comes to our framework token sets aggregation, we formulate the sets of token anchors and unselected tokens as two discrete distributions as:
\begin{equation} 
\label{eq:discrete23}
    U=\sum_{m=1}^{M}u_m\delta_{\bm{X}_m} \hspace{2em} \text{and} \hspace{2em} V=\sum_{n=1}^{N}v_n\delta_{\bm{X}_n},
\end{equation}
where $\bm{u}$ and $\bm{v}$ are the discrete probability vectors that sum to 1, and $\delta_{\bm{f}}$ is a Dirac delta function positioned at support point $\bm{f}$ in the embedding space. Given two support token points $\bm{X}_m$ and $\bm{X}_n$, the cost function is written as $\bm{C} (\bm{X}_m,\bm{X}_n )= 1- \text{sim}(\bm{X}_m,\bm{X}_{n})= 1-\frac{\bm{X}_m^\top\bm{X}_{n}}{||\bm{X}_m||\cdot||\bm{X}_{n}||}$. 
For simplicity, in this discrete situation, $\bm{C} \in \mathbb{R}^{M\times N}$ is a cost matrix in which each point denotes the cost between $\bm{X}_m$ and $\bm{X}_n$.
Then, the total distance of these two distributions is written as:
\begin{equation} 
\label{eq:cost2}
    <\bm{T},\bm{C}> = \sum_{m=1}^{M}\sum_{n=1}^{N}\bm{T}_{m,n}\bm{C}_{m,n},
\end{equation}
where the $\bm{T} \in \mathbb{R}^{M\times N} $ is a matrix of transport plan, which is learned to minimize the total distance. Each point $\bm{T}_{m,n} \in \bm{T}$ is a weight of local cost $\bm{C}_{m,n} $.

The above optimization problem of optimal transport is formulated as:
\begin{equation} 
\label{eq:optimization2}
    \begin{aligned}
    & d_{\text{OT}}(\bm{u},\bm{v}|\bm{C}) = \underset{\bm{T}}{\text{minimize}}
     <\bm{T},\bm{C}> \\
    & \text{subject to}
    ~~~~~~\bm{T}\bm{1}_N = \bm{u},\; \bm{T}^\top \bm{1}_M = \bm{v},\; \bm{T} \in \mathbb{R}^{M \times N}_+.
    \end{aligned}
\end{equation}
These constraints of $\bm{T}$ are used to match its marginal distributions and original discrete distributions in Eq.~\ref{eq:discrete23}. In our framework, we 
consider token anchors $\bm{X}_m$ and unselected tokens $\bm{X}_n$ equally and thus $\bm{u} = \bm{1}_{M\times 1}/M$ and $\bm{v} = \bm{1}_{N\times 1}/N$.

As directly optimizing the above objective is always time-consuming, we apply the Sinkhorn distance~\citep{cuturi2013sinkhorn} to use an entropic constraint for fast optimization.
The optimization problem with a Lagrange multiplier of the entropy constraint is:
\begin{equation} \label{eq:Sinkhorn-app-1}
\begin{aligned}
& d_{\text{OT},\lambda}(\bm{u},\bm{v}|\bm{C})=\underset{\bm{T}}{\text{minimize}}
 <\bm{T},\bm{C}> - \lambda h(\bm{T})\\
& \text{subject to}
~~~~~~~\bm{T}\bm{1}_N = \bm{u},\; \bm{T}^\top \bm{1}_M = \bm{v},\; \bm{T} \in \mathbb{R}^{M \times N}_+,
\end{aligned}
\end{equation}
where $h(\cdot) $ is entropy and $\lambda \geq 0$ is a hyper-parameter. Then we can have a fast optimization solution with a few iterations as:
\begin{equation} \label{eq:Sinkhorn-app-2}
\bm{T}^*= \text{diag}(\bm{u}^{(t)}) \exp(-\bm{C}/\lambda) \text{diag}(\bm{v}^{(t)}),
\end{equation}
where $t$ denotes iteration and in each iteration 
$\bm{u}^{(t)} =\bm{u}/\left((\exp(-\bm{C}/\lambda)\bm{v}^{(t-1)}\right) $ and  $\bm{v}^{(t)} =\bm{v}/\left((\exp(-\bm{C}/\lambda)^\top\bm{u}^{(t)}\right) $, with the initiation $\bm{v}^{(0)} = \bm{1} $. The detailed algorithms of the training and testing processes are shown in Algorithm~\ref{Algorithm1} and ~\ref{Algorithm:inter}

\renewcommand{\algorithmicrequire}{\textbf{Input:}}
\renewcommand{\algorithmicensure}{\textbf{Output:}}
\renewcommand{\algorithmicrequire}{\textbf{Input:}}
\begin{algorithm}[tb]
\caption{\!\!\! \textbf{:} Intra-Frame Inference of AOT with OT}

\label{Algorithm1}
    \begin{algorithmic}[1]
     \Require Testing video with $F$ frames $\bm{X} = \{\bm{x}\}$, image tokens per-frame: $\bm{x}=\{x_1,...,x_N\} \in \mathbb{R}^{N \times d}$ with $M$ token anchors $\mathbf{x}_V^{\texttt{a}} \in \mathbb{R}^{M \times d}$ and $(N-M)$ unselected tokens $\mathbf{x}_V^{\texttt{u}} \in \mathbb{R}^{(N-M) \times d}$ ($M < N$), entropy parameter $\lambda$, maximum number of iterations $\mathbf{Iter}$. 
    \Ensure  Aggregated token anchors of each frame.
    \For {$x=1,2,\dots, \bm{X}$}
        \State{Calculate the cost matrix $\bm{C}=\bm{1}-\bm{{(x}^a_V)}^\top\bm{({x}^{u}_{V})} \in \mathbb{R}^{M \times N}$ of each frame.}
        \State{Calculate the OT distance with a loop:}
        Initialize the $\bm{v}^{(0)} = \bm{1} $, $\delta =0.01$ and $\Delta_{v}= \infty$
        \For{$t_{i}=1,2,\dots,\mathbf{Iter}$}
        \State{Update $\bm{u}^{(t_{i})} =\bm{u}/((\exp(-\bm{C}/\lambda)\bm{v}^{(t_{i}-1)}) $}
        \State{Update $\bm{v}^{(t_{i})} =\bm{v}/((\exp(-\bm{C}/\lambda)^\top\bm{u}^{(t_{i})}) $}
        \State{Update $\Delta_{v} =\sum |\bm{v}^{(t_{i})}- \bm{v}^{(t_{i}-1)}|/N $}
        \If{$\Delta_{v}< \delta$}
        \State{break}
        \EndIf
        \EndFor
        \State{Obtain optimal transport plan  as $\bm{T}^*_{intra}= \text{diag}(\bm{u}^{(t)}) \exp(-\bm{C}/\lambda) \text{diag}(\bm{v}^{(t)}),$}
        \State{Calculate the OT distance $d^{intra}_{\text{OT}} = <\bm{T}^*_{intra},\bm{C} >$}
        \State{Calculate the received transport mass from all unselected tokens for each token anchor with the OT distance $m_j \;=\; \sum_{i=1}^{N-M} T^*_{ij}$}
        \State{Update each token anchor by a mass-normalized OT aggregation from unselected tokens $\tilde{\mathbf{x}}^{a}_j
        \;=\;
        \frac{
            \mathbf{x}^{a}_j
            \;+\;
            \lambda_{intra} \sum_{i=1}^{N-M} T^*_{ij} \,\mathbf{x}^{u}_i
        }{
            1 \;+\; \lambda_{intra} m_j
        }$}
        \State{Update the intra-frame compressed token set $\tilde{\mathbf{x}}_V^{\texttt{a}}
        \;=\;
        \left\{\tilde{\mathbf{x}}^{a}_1, \dots,
        \tilde{\mathbf{x}}^{a}_M\right\}
        \in \mathbb{R}^{M \times d}$}
    \EndFor
    \end{algorithmic}
\end{algorithm}

\renewcommand{\algorithmicrequire}{\textbf{Input:}}
\renewcommand{\algorithmicensure}{\textbf{Output:}}
\begin{algorithm}[tb]
\caption{\!\!\! \textbf{:} Inter-Frame Inference of AOT with OT}
\label{Algorithm:inter}
\begin{algorithmic}[1]
    \Require Video with $F$ frames $\bm{X}$; 
    per-frame intra-frame compressed anchors 
    $\tilde{\bm{X}}^{\texttt{a}}_{V,t} \in \mathbb{R}^{M \times d}$;
    frame-clip partition $\{\mathcal{C}_k\}_{k=1}^{K}$ with 
    $\mathcal{C}_k=\{t_1,\dots,t_L\}$;
    OT entropy $\lambda_{inter}$, max Sinkhorn iters $\mathbf{Iter}$, threshold $\tau$.
    \Ensure  Clip-level token anchors and kept temporal-dynamics tokens.
    \For{$k = 1,2,\dots,K$} \Comment{Process each frame clip $\mathcal{C}_k$}
        \State Initialize clip anchors 
        $\bm{A}^{(1)} \gets \tilde{\bm{X}}^{\texttt{a}}_{V,t_1}$,
        high-change set $\mathcal{D}_k \gets \emptyset$.
        \For{$\ell = 2,3,\dots,L$} \Comment{Subsequent frames}
            \State $\bm{S}^{(\ell)} \gets \tilde{\bm{X}}^{\texttt{a}}_{V,t_\ell}$,
            $\bm{C}^{(\ell)} \gets \bm{1} - \bm{A}^{(\ell-1)} (\bm{S}^{(\ell)})^\top$.
            \State Compute OT plan by Sinkhorn-Knopp:
            \[
                \bm{T}^{(\ell)\,*}_{\text{inter}}
                \gets \mathrm{SinkhornOT}
                \big(\bm{C}^{(\ell)}, \lambda_{inter}, \mathbf{Iter}\big).
            \]
            \State Row-normalize 
            $p^{(\ell)}_{ij} =
            T^{(\ell)\,*}_{ij} /
            \sum_{j'} T^{(\ell)\,*}_{ij'}$, 
            and $q^{(\ell)}_i = \max_j p^{(\ell)}_{ij}$.
            \State Collect high-change tokens:
            $\mathcal{D}_k \gets \mathcal{D}_k 
                \cup \{\bm{s}^{(\ell)}_i \mid q^{(\ell)}_i < \tau\}$.
            \State Update anchors using temporally stable tokens
            ($q^{(\ell)}_i \ge \tau$):
            \[
                \bm{a}^{(\ell)}_j
                =
                \frac{
                    \bm{a}^{(\ell-1)}_j
                    +
                    \lambda_{inter}
                    \sum_{i:\,q^{(\ell)}_i \ge \tau}
                    p^{(\ell)}_{ij}\,\bm{s}^{(\ell)}_i
                }{
                    1
                    +
                    \lambda_{inter}
                    \sum_{i:\,q^{(\ell)}_i \ge \tau}
                    p^{(\ell)}_{ij}
                },\;\; j=1,\dots,M.
            \]
            \State $\bm{A}^{(\ell)} \gets \{\bm{a}^{(\ell)}_j\}_{j=1}^{M}$.
        \EndFor
        \State \textbf{Output} for clip $\mathcal{C}_k$:
        anchors $\bm{A}^{(L)}$ and temporal tokens $\mathcal{D}_k$.
    \EndFor
\end{algorithmic}
\end{algorithm}

\subsection{Optimal Transport and Sinkhorn Iteration}
\label{sec:sinkhorn}

In this subsection, we briefly introduce the derivation of the Sinkhorn-Knopp Iteration~\cite{cuturi2013sinkhorn} algorithm, which we emphasize is not our contribution and belongs to textbook knowledge.
The mathematical formula of the Optimal Transport problem is defined in Eq.~\ref{eq:discrete23} and Eq.~\ref{eq:optimization2}. This is a linear program that can be solved in polynomial time. For dense token sets case, however, the resulting linear program is large, involving the square of feature dimensions with anchors in all scales. This issue can be addressed by a fast iterative solution, which converts the optimization target above into a non-linear but convex form with an entropic regularization term $E$ added:
\begin{alignat}{2}
\begin{split}
\min_{T}\quad &\sum\limits_{i=1}^{m}\sum\limits_{j=1}^{n} C_{ij}T_{ij} + \gamma E(T_{ij}),
\end{split}
\label{Eq2}
\end{alignat}
where $E(T_{ij})=T_{ij} (\log T_{ij}-1)$. $\gamma$ is a constant hyper-parameter controlling the intensity of the regularization term. According to the Lagrange Multiplier Method, the constraint optimization target in Eq.~\ref{Eq2} can be converted to a non-constraint target:
\begin{alignat}{2}
\begin{split}
    \min_{t}\quad & \sum\limits_{i=1}^{m}\sum\limits_{j=1}^{n} C_{ij}T_{ij} + \gamma E(T_{ij}) + \\ 
    & \alpha_j(\sum\limits_{i=1}^{m} T_{ij} - d_j) + \beta_i(\sum\limits_{j=1}^{n} T_{ij} - s_i),
\end{split}
\end{alignat}
where $\alpha_j (j=1,2,...n)$ and $\beta_i (i=1,2,...,m)$ are Lagrange multipliers. 
Nothing that the $i$-th supplier (unselected token) holds $s_i$ units of contexts while the $j$-th demander (token anchor) needs $d_j$ units of contexts. Transporting cost for each unit of context from supplier $i$ to demander $j$ is denoted by $C_{ij}$. The target of the OT problem is to find out a transportation plan $T^{*}=\{T_{i,j} | i=1,2,...m, j=1,2,...n\}$, according to which all contexts from suppliers can be transported to demanders at a minimal transportation cost. 
By letting the derivatives of the optimization target equal to 0, the optimal plan $T^{*}$ is then resolved as:
\begin{equation}
\begin{split}
    T_{ij}^{*} = \exp(-{\frac{\alpha_j}{\gamma}})\exp(-{\frac{C_{ij}}{\gamma}})\exp(-{\frac{\beta_i}{\gamma}}).
\end{split}
\end{equation}

Letting $u_j=\exp(-{\frac{\alpha_j}{\gamma}}), v_i=\exp(-{\frac{\beta_i}{\gamma}}), M_{ij}=\exp(-{\frac{C_{ij}}{\gamma}})$, the following constraints can be enforced:
\begin{alignat}{2}
\sum_i T_{ij}=u_j(\sum_i M_{ij}v_i)=d_j,\\
\sum_j T_{ij}=(u_j\sum_i M_{ij})v_i=s_i.
\end{alignat}

These two equations have to be satisfied simultaneously. One possible solution is to calculate $v_i$ and $u_j$ by repeating the following updating formulas for sufficient steps:
\begin{equation}
\begin{split}
u_j^{t+1}=\frac{d_j}{\sum_iM_{ij}v_i^t},\quad v_i^{t+1}=\frac{s_i}{\sum_jM_{ij}u_j^{t+1}}.
\end{split}
\label{skiter}
\end{equation}
The updating rule in Eq.~\ref{skiter} is also known as the Sinkhorn-Knopp Iteration. After repeating this iteration $t$ times, the approximate optimal plan $T^*$ can be obtained:
\begin{equation}
\begin{split}
T^* = diag(u) M diag(v).
\end{split}\label{optimal_assigning_plan}
\end{equation}

$\gamma$ and $T$ are empirically set to 0.1 and 100.

\section{More Implementation Details}
\label{sec:more_imples}

Our method is implemented on LLaVA-OneVision-7B~\cite{li2024llava} and LLaVA-Video-7B~\cite{zhang2024video} models. Both evaluation and inference use 8 NVIDIA A100 GPUs, with each having 40GB memory availability. 
The number of intra-frame token anchors is set to be 126 for a 10\% token retention budget, 144 for 15\%, 196 for 20\%, and 205 for 25\% when employing LLaVA-OneVision-7B with 32 sampled frames. The number of intra-frame token anchors is set to be 108 for a 10\% token retention budget, 144 for 15\%, 176 for 20\%, and 198 for 25\% when employing LLaVA-Video-7B with 64 sampled frames. Both of these two configurations will adjust the keep ratio for inter-frame compression to match the overall token retention budget by tuning $\tau$, respectively. Both intra- and inter-frame Sinkhorn-Knopp iterations are conducted with 100, and the entropy parameters $\lambda$ are 0.1. We found that optimization iteration makes little impact on the compression performance, therefore setting with number 100 as default.

\section{Dynamic Clustering Frame Clip}
\label{sec:dynamic_clip}

Following FastVID~\cite{shen2025fastvid}, we employ Dynamic Temporal Segmentation, an adaptive and simple method that segments boundaries according to video complexity using global level features. This Dynamic Clustering approach achieves both temporal structure and high intra-segment similarity, generating fewer partitions for simple scenes and finer ones for more complex scenes. Meanwhile, this method mitigates the issue of the fixed length, which preserves temporal order but potentially groups visually dissimilar frames. As shown in Table.~\ref{tab:dynamic_ov_benchmark} and Table.~\ref{tab:dynamic_diff_backbone}, the proposed AOT equipped with Dynamic Clustering to segment the frame clips still consistently leads to competitive and superior performances across various video benchmarks, further demonstrating the practical advantage of our method.

\begin{table*}[t]
\centering
\caption{ 
Comparison of state-of-the-art methods on LLaVA-OneVision~\cite{li2024llava} across video benchmarks. The best performance among those with similar retention ratios is highlighted in \textbf{bold}, while the second best will be denoted as \underline{underlined}. \textbf{AOT} \textit{w} Dyn denotes we apply dynamic temporal segmentation to obtain adaptive frames within each clip, following FastVID~\cite{shen2025fastvid}.
}
\vspace{-0.2cm}
\resizebox{\textwidth}{!}{%
\begin{tabular}{l|ccc|cccc|cc}
\toprule
\multirow{2}{*}{Method} &
\multirow{2}{*}{\makecell{Prefilling\\FLOPs (T) $\downarrow$}} & \multirow{2}{*}{\makecell{FLOPs\\Ratio $\downarrow$}} & \multirow{2}{*}{\makecell{Before LLM\\Retained Ratio}}
& \multirow{2}{*}{\makecell{MVBench\\ $\uparrow$}} & \multirow{2}{*}{\makecell{EgoSchema\\ $\uparrow$}} & \multirow{2}{*}{\makecell{LongVideo\\Bench $\uparrow$}} & \multirow{2}{*}{\makecell{VideoMME\\ $\uparrow$}} & \multicolumn{2}{c}{Avg. $\uparrow$}  \\
&&&&&&&& Score & \% \\
\midrule
\rowcolor{gray!20} 
LLaVA-OV-7B & 40.8 & 100\% & 100\% 
& 58.3 & 60.4 & 56.4 & 58.6 & 58.4 & 100 \\ %
FastV~\cite{chen2024image} & 9.3 & 22.8\% & 100\%
& 55.9 & 57.5 & \textbf{56.7} & 56.1 & 56.5 & 96.7 \\ %
PDrop~\cite{xing2024pyramiddrop} & 10.5 & 25.7\% & 100\%
& 56.1 & 58.0 & 54.1 & 56.4 & 56.2 & 96.2 \\
DyCoke \cite{tao2025dycoke} & 8.7 & 21.3\% & 25\% & 53.1 & 59.5 & 49.5 & 54.3 & 54.1 & 92.6 \\

VisionZip~\cite{yang2025visionzip} & 8.7 & 21.3\% & 25\%
& 57.9 & 60.3 & \underline{56.5} & \textbf{58.2} & \underline{58.2} & \underline{99.7} \\ %

PruneVid~\cite{huang2024prunevid} & 8.7 & 21.3\% & 25\%
& 57.4 & 59.9 & 55.7 & 57.4 & 57.6 & 98.6 \\ %

FastVID~\cite{shen2025fastvid} & 8.7 & 21.3\% & 25\%
& 56.5 & - & 56.3 & \underline{58.0} & - & - \\

\textbf{AOT} & 8.7 & 21.3\% & 25\%
& \textbf{58.7} & \underline{61.3} & 56.3 & 57.5 & \textbf{58.5} &  \textbf{100.0} \\ %

\textbf{AOT} \textit{w} Dyn & 8.7 & 21.3\% & 25\%
& \underline{58.5} & \textbf{61.7} & 56.3 & 57.5 & \textbf{58.5} &  \textbf{100.0} \\ %

\midrule
VisionZip~\cite{yang2025visionzip} & 7.0 & 17.2\% & 20\%
& 57.7 & 59.8 & 55.2 & \textbf{57.9} & 57.7 & 98.8 \\ %

PruneVid~\cite{huang2024prunevid} & 7.0 & 17.2\% & 20\%
& 57.2 & 59.7 & 54.7 & 56.9 & 57.1 & 97.8 \\ %

FastVID~\cite{shen2025fastvid} & 7.0 & 17.2\% & 20\%
& 56.3 & - & \textbf{57.1} & \textbf{57.9} & - & - \\

\textbf{AOT} & 7.0 & 17.2\% & 20\%
& \underline{58.1}  & \underline{61.3} & \underline{56.2} & \underline{57.2} & \textbf{58.2} & \textbf{99.7} \\ %

\textbf{AOT} \textit{w} Dyn & 7.0 & 17.2\% & 20\%
& \textbf{58.2}  & \textbf{61.5} & 55.9 & 56.8 & \underline{58.1} & \underline{99.5}\\ %

\midrule
VisionZip~\cite{yang2025visionzip} & 5.2 & 12.7\% & 15\%
& 56.5 & 59.8 & 54.4 & 56.1 & 56.7 & 97.1 \\ %

PruneVid~\cite{huang2024prunevid} & 5.2 & 12.7\% & 15\%
& \underline{56.8} & 59.7 & \underline{55.4} & \underline{56.6} & 57.1 & 97.8 \\ %

FastVID~\cite{shen2025fastvid} & 5.2 & 12.7\% & 15\% & 56.0 & - & \textbf{56.2} & \textbf{57.7} & - & - \\

\textbf{AOT} & 3.4 & 8.3\% & 15\%
& \textbf{57.8}  & \textbf{61.3} & 55.2 & \underline{56.6} & \textbf{57.7} & \textbf{98.8} \\ %

\textbf{AOT} \textit{w} Dyn & 5.2 & 12.7\% & 15\%
& \textbf{57.8}  & \underline{61.2} & 55.1 & 56.1 & \underline{57.6} & \underline{98.6} \\ %

\midrule
VisionZip~\cite{yang2025visionzip} & 3.4 & 8.3\% & 10\% 
& 53.5 & 58.0 & 49.3 & 53.4 & 53.5 & 91.6 \\ %

PruneVid~\cite{huang2024prunevid} & 3.4 & 8.3\% & 10\% 
 & 56.2 & 59.8 & \underline{54.5} & 56.0 & \underline{56.6} & \underline{96.9} \\ %

FastVID~\cite{shen2025fastvid} & 3.4 & 8.3\% & 10\% 
& 55.9 & - & \textbf{56.3} & \textbf{57.3} & - & - \\

\textbf{AOT} & 5.2 & 12.7\% & 10\%
& \underline{57.0}  & \textbf{60.6} & 54.2 & \underline{56.1} & \textbf{57.0} & \textbf{97.6} \\ %

\textbf{AOT} \textit{w} Dyn & 3.4 & 8.3\% & 10\%
& \textbf{57.2}  & \underline{60.0} & 53.1 & 55.7 & 56.5 & 96.7 \\ %

\bottomrule

\end{tabular}
}
\label{tab:dynamic_ov_benchmark}
\vspace{-0.2cm}
\end{table*}

\begin{table*}[t]
\centering
\caption{Comparison of state-of-the-art methods on LLaVA-Video~\cite{zhang2024video} across video benchmarks. The best performance among those is highlighted in \textbf{bold}, while the second best will be denoted as \underline{underlined}, demonstrating consistent effectiveness. \textbf{AOT} \textit{w} Dyn denotes we apply dynamic temporal segmentation to obtain adaptive frames within each clip, following FastVID~\cite{shen2025fastvid}.
}
\vspace{-0.2cm}
\resizebox{\textwidth}{!}{%
\begin{tabular}{l|ccc|cccc|cc}
\toprule
\multirow{2}{*}{Method} &
\multirow{2}{*}{\makecell{Prefilling\\FLOPs (T) $\downarrow$}} & \multirow{2}{*}{\makecell{FLOPs\\Ratio $\downarrow$}} & \multirow{2}{*}{\makecell{Before LLM\\Retained Ratio}}
& \multirow{2}{*}{\makecell{MVBench\\ $\uparrow$}} & \multirow{2}{*}{\makecell{EgoSchema\\ $\uparrow$}} & \multirow{2}{*}{\makecell{LongVideo\\Bench $\uparrow$}} & \multirow{2}{*}{\makecell{VideoMME\\ $\uparrow$}} & \multicolumn{2}{c}{Avg. $\uparrow$}  \\
&&&&&&&& Score & \% \\
\midrule
 \cellcolor{gray!20}{LLaVA-Video-7B} & \cellcolor{gray!20}{80.2} & \cellcolor{gray!20}{100\%} & \cellcolor{gray!20}{100\%} 
 & \cellcolor{gray!20}{60.4} & \cellcolor{gray!20}{57.2} & \cellcolor{gray!20}{58.9} & \cellcolor{gray!20}{64.3} & \cellcolor{gray!20}{60.2} & \cellcolor{gray!20}{100} \\ %

 FastV~\cite{chen2024image} & 17.1 & 21.3\% & 100\% 
& 54.3 & 54.1 & 55.0 & 58.8 & 55.6 & 92.4 \\ %

 PDrop~\cite{xing2024pyramiddrop} & 19.5 & 24.3\% & 100\% 
& 55.9 & 54.3 & 54.7 & 61.9 & 56.7 & 94.2 \\ %

 VisionZip~\cite{yang2025visionzip} & 9.3 & 18.9\% & 25\%
& 56.7 & \underline{54.7} & 54.7 & 60.7 & 56.7 & 94.2 \\ %

 DyCoke~\cite{yang2025visionzip} & 9.3 & 18.9\% & 25\%
& 50.8 & - & 53.0 & 56.9 & - & - \\ %

 \textbf{AOT} & 9.3 & 18.9\% & 25\%
& \underline{58.8}  & \textbf{55.4} & \underline{56.2} & \underline{62.4} & \underline{58.2} & \underline{96.7} \\ %

 \textbf{AOT} \textit{w} Dyn & 9.3 & 18.9\% & 25\%
& \textbf{59.1}  & \textbf{55.4} & \textbf{57.1} & \textbf{62.6} & \textbf{58.6} & \textbf{97.3} \\ %

\hline

 VisionZip~\cite{yang2025visionzip} & 9.3 & 11.6\% & 15\%
& 56.7 & \underline{54.7} & 54.7 & 60.7 & 56.7 & 94.2 \\ %

 \textbf{AOT} & 9.3 & 11.6\% & 15\%
& \textbf{57.8}  & \textbf{55.2} & \underline{55.0} & \textbf{62.0} & \textbf{57.5} & \textbf{95.5} \\ %

 \textbf{AOT} \textit{w} Dyn & 9.3 & 11.6\% & 15\%
& \underline{57.6}  & 54.6 & \textbf{55.3} & \underline{61.3} & \underline{57.2} & \underline{95.0} \\ %

\bottomrule

\end{tabular}
}%
\label{tab:dynamic_diff_backbone}
\vspace{-0.4cm}
\end{table*}

\begin{table}[tb]
\centering
\caption{Ablation: the impact of random token selection.}
\vspace{-0.2cm}
\resizebox{0.48\textwidth}{!}{%
\begin{tabular}{l|cccc|cc}
\toprule
\multirow{1}{*}{Method} & \multirow{1}{*}{\makecell{MVBench}} & \multirow{1}{*}{\makecell{EgoSchema}} & \multirow{1}{*}{\makecell{LongVideo\\Bench}} & \multirow{1}{*}{\makecell{VideoMME}} & \multicolumn{2}{c}{Avg.}  \\
&&&&& Score~$\uparrow$ & \%~$\uparrow$ \\
\midrule
\rowcolor{gray!20} 
Vanilla & 58.3 & 60.4 & 56.4 & 58.6 & 58.4 & 100 \\ %

\hline

\textit{w} Random Anchors & 55.1 & 59.3 & 52.6 & 53.3 & 55.1 & 94.3 \\ %

AOT & 57.0 & 60.6 & 54.2 & 56.1 & 57.0 & 97.6 \\ %

\bottomrule
\end{tabular}
}
\label{tab:ablations_random}
\vspace{-0.4cm}
\end{table}

\section{Random Token Anchors Selection Ablation}
\label{sec:random}

In Table.~\ref{tab:ablations_random}, we further conduct more ablation study by randomly selecting the token anchors within intra-frame level to demonstrate the importance of high-quality token anchors establishment initially, while randomly selecting noisy tokens deteriorates performances drastically.

\section{More Visualizations}
\label{sec:visualization}

\begin{figure*}[t]
    \centering
    \includegraphics[width=\linewidth]{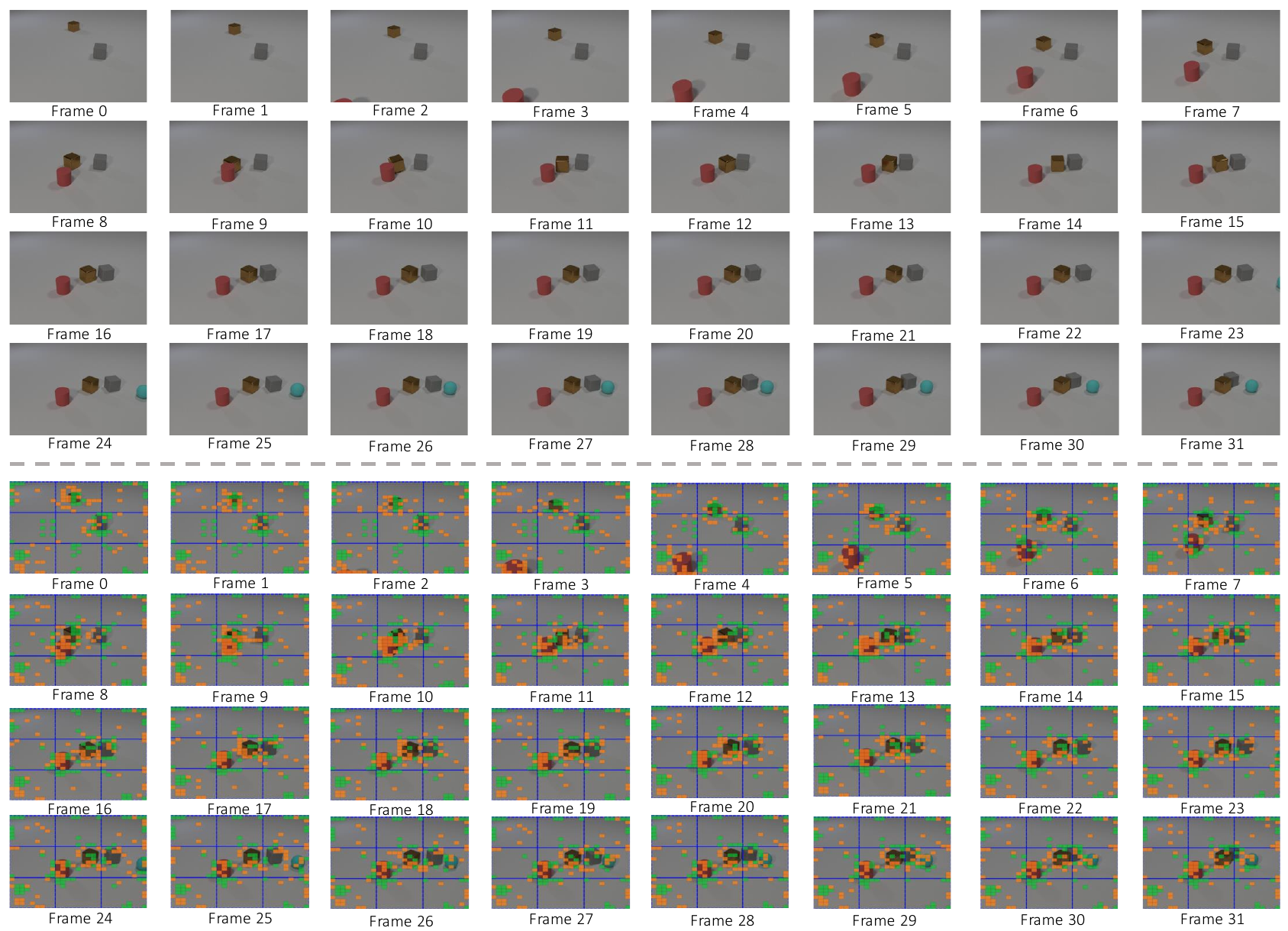}
    \caption{Qualitative visualizations of our \textcolor{green}{Local}-\textcolor{orange}{Global} token anchors evolution across consecutive frames on MVBench sample while optimal transport is adopted to aggregate necessary information from unselected tokens to help LLM precess better. The top is the original sampled frames while the bottom is the corresponding tokens visualization.}
\label{fig:supple_vis_01}
\end{figure*}

\begin{figure*}[t]
    \centering
    \includegraphics[width=\linewidth]{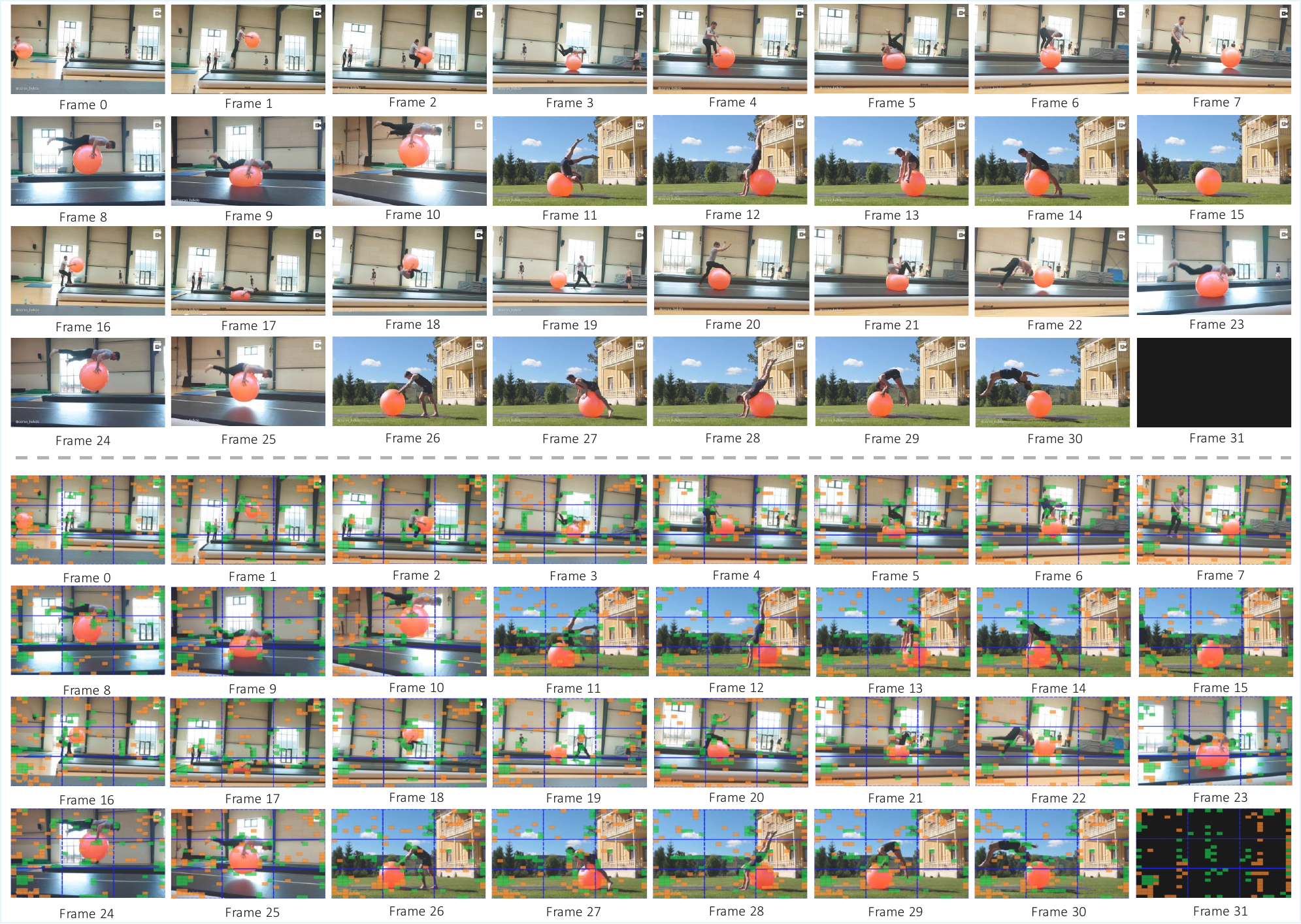}
    \caption{Qualitative visualizations of our \textcolor{green}{Local}-\textcolor{orange}{Global} token anchors evolution across consecutive frames on VideoMME sample while optimal transport is adopted to aggregate necessary information from unselected tokens to help LLM precess better. The top is the original sampled frames while the bottom is the corresponding tokens visualization.}
\label{fig:supple_vis_02}
\end{figure*}

As shown in Fig.~\ref{fig:supple_vis_01} and Fig.~\ref{fig:supple_vis_02}, we visualize the selected token anchors initially. Since both LLaVA-OneVision-7B~\cite{li2024llava} and LLaVA-Video-7B~\cite{zhang2024video} utilize spatial pooling for the neighboring tokens, which makes the visualization for the temporal token marking difficult, we opt to showcase the initial intra-frame token anchors across the consecutive sampled frames to help better illustrate the effectiveness of optimal transport in this paper to aggregate necessary information from unselected tokens to help LLM process better.

\section{Limitation and Future Works}
\label{sec:limit_future}
\noindent \textbf{Limitation.}
This paper focuses on aggregating informative semantics and context into a compact set of remaining tokens via OT-based aggregation, rather than naively discarding or averaging them.
However, the inter-frame OT pruning module is still heuristic, as there is no principled way to construct high-quality temporal token anchors, unlike the intra-frame case where single images are encoded by powerful visual encoders~\cite{radford2021learning,zhai2023sigmoid}.
Although our method supports both fixed and dynamical temporal segmentation frame clips, temporal boundaries are still noisy, so visually dissimilar frames may be grouped within the same clip, degrading performance in complex video scenarios.

\noindent \textbf{Future Works.}
While we use OT in a training-free manner to aggregate local–global token anchors across intra- and inter-frame levels, the whole inference flow is still end-to-end differentiable, since the transport plan is computed by a small number of matrix multiplications in the forward pass. Thus, gradients can still be back-propagated through the OT optimization strategy, enabling the entire system (including the iterative updates) fully differentiable. First, It is therefore promising to explore model fine-tuning or instruction tuning with OT, aiming for a more competitive and efficient token reduction framework. Second, beyond efficiency, OT offers a principled way to impose structure and consistency across time, which suggests leveraging auxiliary supervision signals to enhance long-context reasoning and extend Video LLMs toward 3D/4D spatial intelligence. In particular, incorporating signals such as depth, camera motion, correspondence, 3D grounding, spatiotemporal consistency, or trajectory-aware cues could encourage the aggregated anchors to preserve geometry-aware and motion-aware information rather than merely appearance-level semantics. Combining OT-based token aggregation with these auxiliary objectives may improve the model’s ability to maintain persistent world representations, support temporally coherent reasoning, and bridge video understanding with embodied or spatial tasks. Exploring these directions~\cite{li2022expansion,zheng2025video,liu2024less,li2025cross,wang2025ross3d,li2025orthogonal,zhu2024llava,mei2025perla,wang2024uvmap,nie2024t2td} could lead to a unified framework that is not only computationally efficient but also better aligned with the demands of 3D/4D perception and decision-making.

\clearpage
\clearpage

{
    \small
    \bibliographystyle{ieeenat_fullname}
    \bibliography{main}
}


\end{document}